%% file: neurips_2025.tex
\documentclass{article}

\usepackage[preprint]{neurips_2025}

\usepackage[utf8]{inputenc} 
\usepackage[T1]{fontenc}    
\usepackage{hyperref}       
\usepackage{url}            
\usepackage{booktabs}       
\usepackage{amsfonts}       
\usepackage{nicefrac}       
\usepackage{microtype}      
\usepackage{xcolor}         
\usepackage{graphicx}
\usepackage{subfigure}
\usepackage{multirow}
\usepackage{natbib}
\usepackage{algorithm}
\usepackage{algpseudocode}
\usepackage{amsmath,amsfonts,bm}
\usepackage{amsthm}
\usepackage{subcaption}
\usepackage{makecell}
\usepackage{comment}
\usepackage{tcolorbox}
\usepackage{xltabular}
\usepackage[utf8]{inputenc} 
\usepackage{textcomp}

\algdef{SE}[REPEATN]{RepeatN}{End}[1]{\algorithmicrepeat\ #1 \textbf{times}}{\algorithmicend}

\input{math_commands.tex}

\title{Data Efficacy for Language Model Training}

\author{\textbf{Yalun Dai}\;\;\;\textbf{Yangyu Huang}\thanks{Corresponding author}\;\;\;\;\textbf{Xin Zhang}\;\;\;\textbf{Wenshan Wu} \\ 
\;\;\;\textbf{Chong Li}\;\;\; \textbf{Wenhui Lu}\;\;\;\textbf{Shijie Cao}\;\;\;\textbf{Li Dong}\;\;\;\textbf{Scarlett Li}
\\
\vspace{0.2cm}
Microsoft Research
\vspace{-0.3cm}}

\begin{document}

\maketitle

\vspace{-10px}

\begin{abstract}
Data is fundamental to the training of language models (LM).
Recent research has been dedicated to data efficiency, which aims to maximize performance by selecting a minimal or optimal subset of training data.
Techniques such as data filtering, sampling, and selection play a crucial role in this area.
To complement it, we define \textbf{Data Efficacy}, which focuses on maximizing performance by optimizing the organization of training data and remains relatively underexplored.
This work introduces a general paradigm, \textbf{DELT}, for considering \textbf{d}ata \textbf{e}fficacy in \textbf{L}M \textbf{t}raining, which highlights the significance of training data organization.
DELT comprises three components: Data Scoring, Data Selection, and Data Ordering.
Data Scoring assigns a score for each data sample based on its properties, such as quality, difficulty, and learnability.
Data Selection optionally selects a subset from the original training data based on the scores.
Data Ordering utilizes these scores to organize the training data in a new, optimized order, rather than the traditional random shuffling.
Furthermore, we design Learnability-Quality Scoring (LQS), as a new instance of Data Scoring, which considers both the learnability and quality of each data sample from the gradient consistency perspective.
We also devise Folding Ordering (FO), as a novel instance of Data Ordering, which addresses issues such as model forgetting and data distribution bias.
Comprehensive experiments validate the data efficacy in LM training, which demonstrates the following:
Firstly, different DELT instances enhance LM performance to varying degrees without increasing the data scale and model size.
Secondly, among these instances, the combination of our proposed LQS for data scoring and FO for data ordering achieves the most significant improvement.
Lastly, data efficacy can be achieved together with data efficiency by applying data selection.
Therefore, we believe that data efficacy is a promising foundational area in LM training.
\href{https://github.com/microsoft/DELT}{The Code} is publicly available now.
\end{abstract}
\vspace{-10px}

\begin{figure}[!h]
\vspace{-10px}
  \centering
  \includegraphics[width=0.8\columnwidth]
{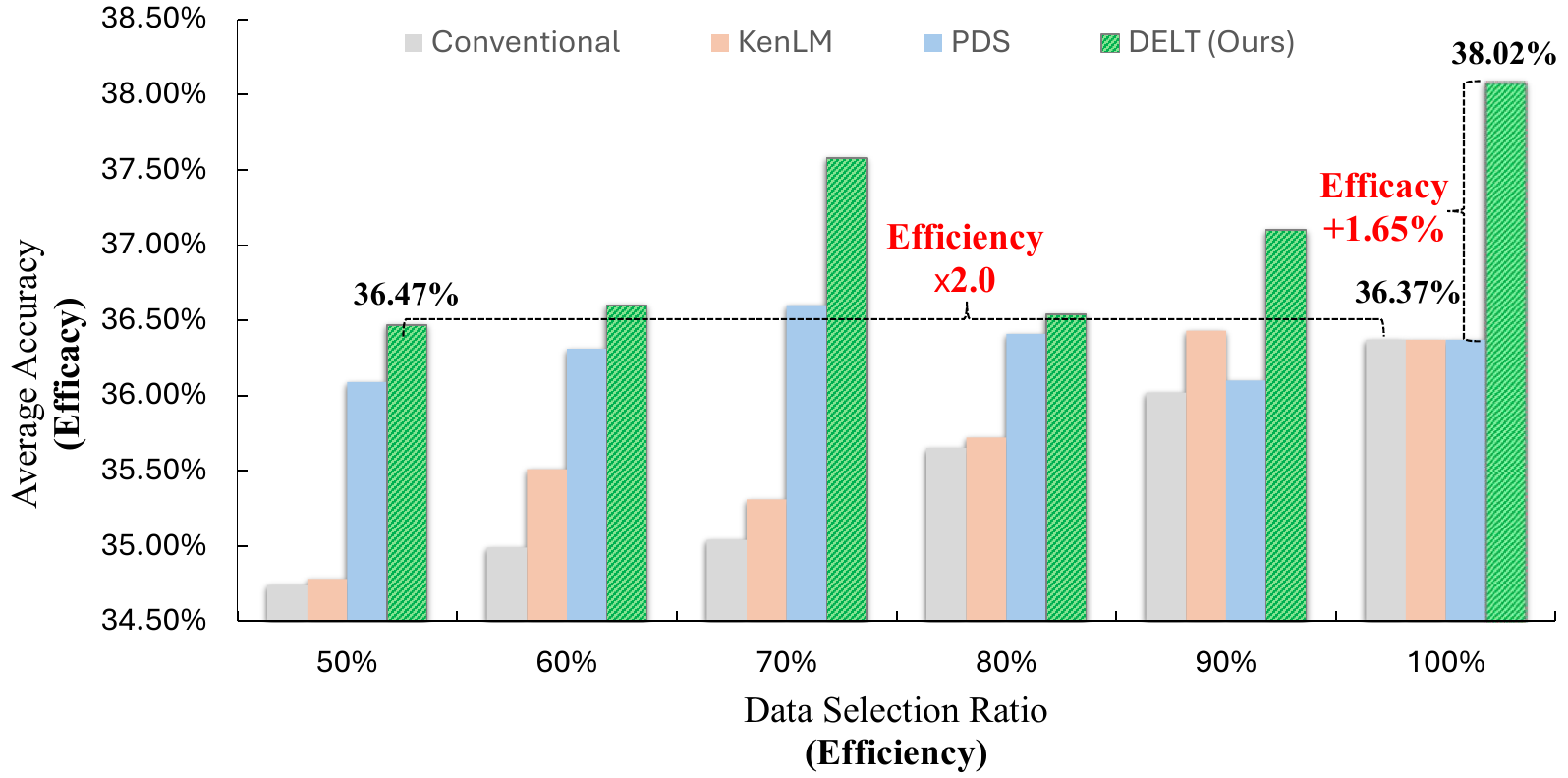}
\vspace{-5px}
  \caption{
  Average result across 8 benchmarks for different methods.
  High performance at the same selection ratio indicates high efficacy, while achieving similar performance with a smaller selection ratio demonstrates high efficiency.
  Our method excels in both efficacy and efficiency.
  }
  \label{fig:cover}
\vspace{-20px}
\end{figure}

\section{Introduction}

The significance of language models \citep{ouyang2022training, achiam2023gpt, dubey2024llama} is immense in modern computational applications.
From natural language processing tasks such as translation \citep{hirschberg2015advances} and sentiment analysis \citep{gunasekaran2023exploring} to more complex applications like automated reasoning \citep{yu2024natural} and conversational agents \citep{kusal2022ai}, language models have revolutionized the way machines understand, generate, and interact with human beings using natural language.
To empower language models having these abilities, data is central to their training and serves as the foundation from which models learn knowledge based on linguistic patterns and structures.
Consequently, meticulous data curation is essential to ensure consistently high model performance across various applications.

Recent research has therefore concentrated on \textbf{data efficiency}, \textit{selecting the smallest or highest-quality subset of the corpus that still yields strong results} \citep{albalak2024dssurvey, xie2023data, gu2024data}.
Once that subset is chosen, however, every surviving sample is ordinarily treated the same, and the order in which samples are shown to the model is random.
In this work, we define \textbf{data efficacy} as \textit{improving model performance by optimizing the organization of training data}.
This area complements data efficacy and is still in its early stage, with its potential demonstrated through curriculum learning \citep{campos2021curriculum, wang2021survey} that feeds examples to model from easy to hard.



In this context, we notice that the latest generation of language models \citep{openAI4o2024, dubey2024llama, team2023gemini} typically trains for only a few epochs, usually just one, due to the vast scale of training datasets but limited computing power.
These models contrast with previous generations \citep{cho2014learning, hochreiter1997long} by the scaling law \citep{kaplan2020scaling}, which trained over many epochs and often led to overfitting.
This aligns with the findings of QQT \citep{goyal2024scaling}, which shows that high-quality data quickly loses its utility after being used repeatedly.
In other words, it is more effective to utilize a large amount of training data with few epochs rather than rely on high-quality data with many epochs.
\textbf{Consequently, effectively organizing the training dataset is essential for enhancing the performance of language models trained with only a few epochs}.

Expanding on this insight, we propose a general paradigm for data efficacy that achieves benefits without altering the dataset content and the model architecture, which makes it an almost cost-free approach.
Specifically, this paradigm incorporates data scoring, data selection, and data ordering components.
Data scoring assigns a score to each sample, which reflects factors like difficulty, quality, diversity, and learnability.
Data selection involves optionally choosing a subset of the original training data based on these scores.
Data ordering then organizes the training data according to these scores, either in ascending, descending, or other arrangements.
Curriculum learning \citep{campos2021curriculum, wang2021survey} can be viewed as a specific example within our paradigm, with ascending ordering based on difficulty scoring.

To verify the proposed paradigm, we integrate some baseline methods into it and also design new methods respectively for data scoring and ordering.
The key results from Figure \ref{fig:cover} highlight that the proposed DELT significantly improves data efficacy in LM training on a set of typical benchmarks.
Meanwhile, it outperforms existing methods \citep{gu2024data, heafield-2011-kenlm} in data efficiency that further boosts LM performance across all selection ratios.


The main contributions of this paper are as below:

\begin{itemize}
  \item We identified the potential of the underexplored area, data efficacy, in language model training and proposed a general paradigm for this area, DELT, which consists of data scoring, data selection, and data ordering.
  \item We designed an innovative method for data scoring, called Learnability-Quality Scoring (LQS), which evaluates the score for each data sample based on learnability and quality from the gradient consistency perspective.
  \item We devised a novel method for data ordering, named Folding Ordering (FO), which optimizes LM training and mitigates the issues of model forgetting and data distribution bias.
  \item We conducted comprehensive experiments to validate the DELT paradigm on mainstream benchmarks, employing different data scoring and data ordering methods. All instances of the proposed paradigm improved performance, with our design outperforming the rest.
\end{itemize}

Through these contributions, we aim to provide a general paradigm for understanding and applying data efficacy in LM training, paving the way for more effective model development practices.

\newpage


\section{Related Work}

\subsection{Data Sources}
Data source of LM training \citep{anthropic2024,yang2007paml,dubey2024llama,ouyang2022training,team2023gemini} can primarily be categorized into five types: internet data \citep{cc:Rana:2010:Common-Crawl-open-web-scale-crawl}, books \citep{Gutenberg2004}, synthetic data \citep{nikolenko2021synthetic}, physical sensors \citep{kabadayi2006virtual}, and human perception of the real world.
Internet data is the primary source for language model training due to its vast scale.
Books and synthetic data offer high quality, but are limited in scale.
Data from physical sensors and human perception are in other modalities or still under development.
Several studies focus on extracting high-quality datasets for LM training, like C4 \citep{raffel2020exploring}, RefinedWeb \citep{penedo2023refinedweb}, RedPajama \citep{weber2024redpajama}, and RedStone \citep{chang2024redstone}.
All of them utilize an identical data source, CommonCrawl \citep{cc:Rana:2010:Common-Crawl-open-web-scale-crawl}, which captures snapshots of web pages from the entire internet at different periods and contains over 200 billion samples to date.

\subsection{Data Efficiency}
Data efficiency \citep{heafield-2011-kenlm, gu2024data, xie2023data, albalak2024dssurvey} focuses on selecting the most relevant data points for inclusion in training dataset and optimizing the performance of the language model.
This area includes well-researched strategies such as data selection  \citep{heafield-2011-kenlm, gu2024data, yu2024mates,daitfdp}, sampling \citep{xie2023data}, denoising \citep{zhao2021p, hu2021p}, and deduplication \citep{abbas2023semdedup, tirumala2023d4}, all of which aim to select optimal data for efficient model training.
The KenLM \citep{heafield-2011-kenlm} trains a fast and small model for perplexity estimation and treats the perplexity as the data difficulty for LM.
The PDS \citep{gu2024data} evaluates the quality of data samples by measuring the consistency of each sample's gradient direction with a reference direction.
The DSIR \citep{xie2023data} develops an importance weight estimator to select a subset of raw data that mirrors the distribution of the target in a specific feature space.
The MATES \citep{yu2024mates} presents a data influence model that continuously adapts to the evolving data preferences of the pre-trained model, selecting the most effective data for the current stage of pre-training.
The SemDeDup \citep{abbas2023semdedup} utilizes embeddings from pre-trained models to identify and remove data pairs that are semantically similar but not exactly identical.
All these methods develop strategies to decide whether a sample should be retained or discarded. However, for retained samples, language models train on them equally, without considering differences in criteria.

\subsection{Data Efficacy}
Data efficacy, distinct from data efficiency, aims to maximize the performance of language models by optimizing the organization of training data.
Curriculum learning, as described by \citep{campos2021curriculum}, involves starting with simpler examples and progressively tackling more complex ones, aiding in smoother model convergence.
Within curriculum learning, \citep{kim2024strategic} presents an attention score to determine the prompt difficulty, and \citep{chang2021does} introduces a soft edit distance to measure sample difficulty.
Similarly, annealing learning, as outlined by \citep{dubey2024llama}, seeks to improve model performance by initially training on a large, noisy dataset and concluding with a small, high-quality dataset.
All these methods sort training data directly by difficulty or quality.
However, since limited research on data efficacy, there is no established paradigm for effectively organizing training data.

To conclude, data is essential for training language models, and numerous large-scale data sources originate from the internet.
Nevertheless, obtaining incremental public data has become challenging due to the slow growth of CommonCrawl \citep{cc:Rana:2010:Common-Crawl-open-web-scale-crawl} snapshots and the increasing presence of AI-generated content online.
As language models scale up, effectively leveraging existing data sources becomes vital, which makes data efficacy increasingly important.
Despite this, few studies focus on data efficacy in language model training.
To address this gap, we propose a general paradigm for it, where curriculum learning and annealing learning are two specific instances.



\section{Paradigm}

\subsection{Problem Formulation}
Language model (LM) is represented by parameters $\vtheta \in \sR^N$, which is pre-trained from scratch or finetuned from weights on a dataset $\Dtrn = {\{\xtrn_n\}}_{n=1}^{|\Dtrn|}$ with $T$ training steps.
The $N$ is the quantity of $\vtheta$ and $x_n$ is the $n$-th sample in $\Dtrn$.
The goal of the problem is to optimize the LM performance by reorganizing training dataset $\Dtrn$ on downstream tasks, which is named \textbf{Data Efficacy} in this paper.

\begin{figure}[!ht]
  \centering
  \includegraphics[width=1.0\columnwidth]{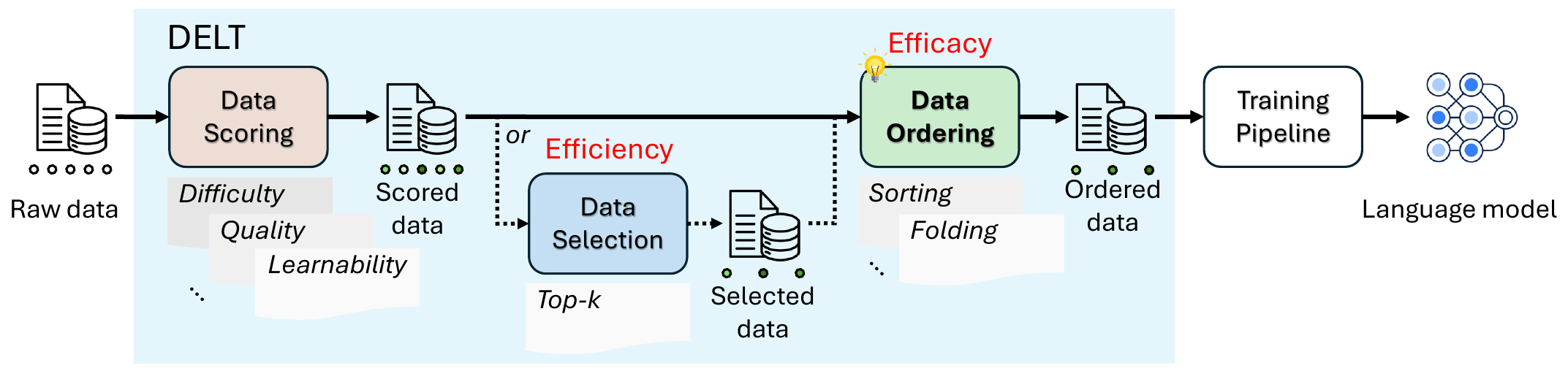}
  \caption{
  Paradigm of Data Efficacy for LM training.
  The blue box represents the paradigm DELT.
  Both the methods for data scoring and data ordering components in DELT are flexible and can be adjusted, including those outlined in Section \ref{sec:method}.
  Meanwhile, the data selection is an optional component that can further improve data efficiency.
  Within the proposed DELT, data efficacy and efficiency are seamlessly compatible, working together to optimize language model performance.
  }
  \label{fig:paradigm}
\end{figure}

\subsection{Paradigm Definition}
A paradigm is proposed in Figure \ref{fig:paradigm} to improve \textbf{data efficacy} in LM training without altering the data content $\Dtrn$ and model parameters $\vtheta$.
It comprises three components:

\begin{itemize}
\item
\textbf{Data Scoring}: it aims to assign a score for each training sample based on specific criteria, such as quality, difficulty, diversity, and learnability. These scores are then applied to guide data selection and data ordering in subsequent stages.
\item
\textbf{Data Selection}: it seeks to select an optimal subset $\Dselect$ from the dataset $\Dtrn$, ensuring that LMs trained on $\Dselect$ achieve the best possible performance.
This process alters the quantity of dataset $\Dtrn$ but does not influence the organization of $\Dselect$.
\item
\textbf{Data Ordering}: it targets at reorganizing the order of training samples in $\Dtrn$ (or $\Dselect$) to create $\Dorder$, such that LMs trained on $\Dorder$ achieve superior performance.
This process focuses on the organization of dataset $\Dtrn$ (or $\Dselect$), while it does not change the dataset scale.
\end{itemize}

Unlike the baseline method, where the language model is trained directly on the raw data $\Dtrn$, and the data efficiency methods that use a selected subset $\Dselect$, DELT processes the raw data $\Dtrn$ as follows:

Firstly, data scoring, defined as $f$, assigns a score vector $\vga$ to the raw data $\Dtrn$, where $\vga$ lies in a $|\Dtrn|$-dimensional simplex.
Samples with large $\vga$ are considered good according to their criteria.
\begin{equation}
\vga = f(\Dtrn) = \left[\gamma_1, \gamma_2, \cdots, \gamma_{|\Dtrn|}\right]^\top
\label{eq:gamma}
\end{equation}
Then, data selection, denoted as $f_{s}$, identifies a subset $\Dselect$ from $\Dtrn$ based on the scores $\vga$ by the selection ratio $r$.
The number of samples $K$ to be selected is determined by $r$.
The function $rank$ provides the ranking index of each element in the set $\vga$ in ascending order.
\begin{equation}
\Dselect = f_{s}(\Dtrn; \vga, K) = \left\{ \xtrn_k \ \middle| \ rank(\vga_{k}) > {|\Dtrn|} - K \text{ and } 1 \le k \le {|\Dtrn|} \right\}
\label{eq:f_s}
\end{equation}
\begin{equation}
K = \lfloor r \cdot |\Dtrn| \rfloor
\label{eq:K}
\end{equation}
Finally, data ordering, represented by $f_{o}$, reorganizes the $\Dtrn$ or $\Dselect$ into a new dataset $\Dtrn'$ with unchanged size, based on a permutation $\pi$ determined by $\vga$.
It could be $\pi_{sort}$ that returns the indices of each element in $\vga$ after sorting or other functions.
\begin{equation}
\Dtrn' = f_{o}(\Dtrn; \vga) = \left[ \xtrn_{\pi(\vga)_{1}}, \xtrn_{\pi(\vga)_{2}}, \cdots, \xtrn_{\pi(\vga)_{|\Dtrn|}} \right]
\label{eq:f_o}
\end{equation}

\textbf{Compatibility of data efficacy and data efficiency in DELT.}
As shown in Figure \ref{fig:paradigm}, the DELT paradigm can build upon data scoring and data ordering by incorporating data selection to further enhance \textbf{data efficiency}. 
The entire DELT process can be defined as a transformation of the original dataset $\Dtrn$ into a reordered dataset $\Dtrn'$:  
\begin{equation}
\Dtrn' = f_{o}(\vga_{o}) \circ f_{s}(\Dtrn; \vga_{s}, K),
\label{eq:f_so}
\end{equation}
where the symbol $\circ$ denotes functional composition.
$\vga_{o}$ and $\vga_{s}$ are the score vectors for data ordering and data selection, respectively.
Since data scoring often requires substantial computation time, both data selection and data ordering in DELT apply a shared score vector for practicality, i.e., $\vga_{o} = \vga_{s} = \vga$.
This process ensures that the most qualified samples are selected and then optimally ordered, thereby significantly improving model performance in both data efficacy and efficiency.


\section{Method}
\label{sec:method}

The DELT paradigm can adopt various specific data scoring and data ordering methods.
Below, we introduce some baseline methods for each of them and propose an improved approach.
These methods are not exhaustive, and other options are also viable.
For data selection, top-K is applied.

\subsection{Data Scoring}

\subsubsection{Baseline Methods}
Existing methods typically focus on attributes such as quality \cite{gu2024data}, difficulty \citep{heafield-2011-kenlm}, noisiness \citep{zhao2021p, hu2021p}, or diversity \citep{abbas2023semdedup, tirumala2023d4} to compute scores for data selection.

\textbf{KenLM} \citep{heafield-2011-kenlm} as a small n-grams model is applied to learn the perplexity of training samples, which is considered to score data samples by their difficulty.

\textbf{PDS} \citep{gu2024data} trains a compact neural network to evaluate the quality of data samples, which is represented by the consistency of each sample’s gradient direction with a reference direction.

However, these methods, designed primarily for data selection, often focus solely on how \textit{good} a sample is, while overlooking the question of \textit{where} a sample contributes most effectively within the context of the entire dataset.

\subsubsection{Our Method}
\label{sec:lqs method}
\textbf{Learnability-Quality Scoring (LQS)} is introduced to address this limitation and make the scorer more attentive to the utility of each data sample.
By incorporating both \textit{learnability} and \textit{quality}, LQS is not only sensitive to low-quality samples but also better weights the impact of samples during model training.
Our method dynamically evaluates how each sample contributes to reducing the downstream loss $J(\vtheta)$ by considering its behavior at different training stages. 

\begin{figure}[!hb]
\vspace{-0.3cm}
  \centering
  \includegraphics[width=1.0\columnwidth]{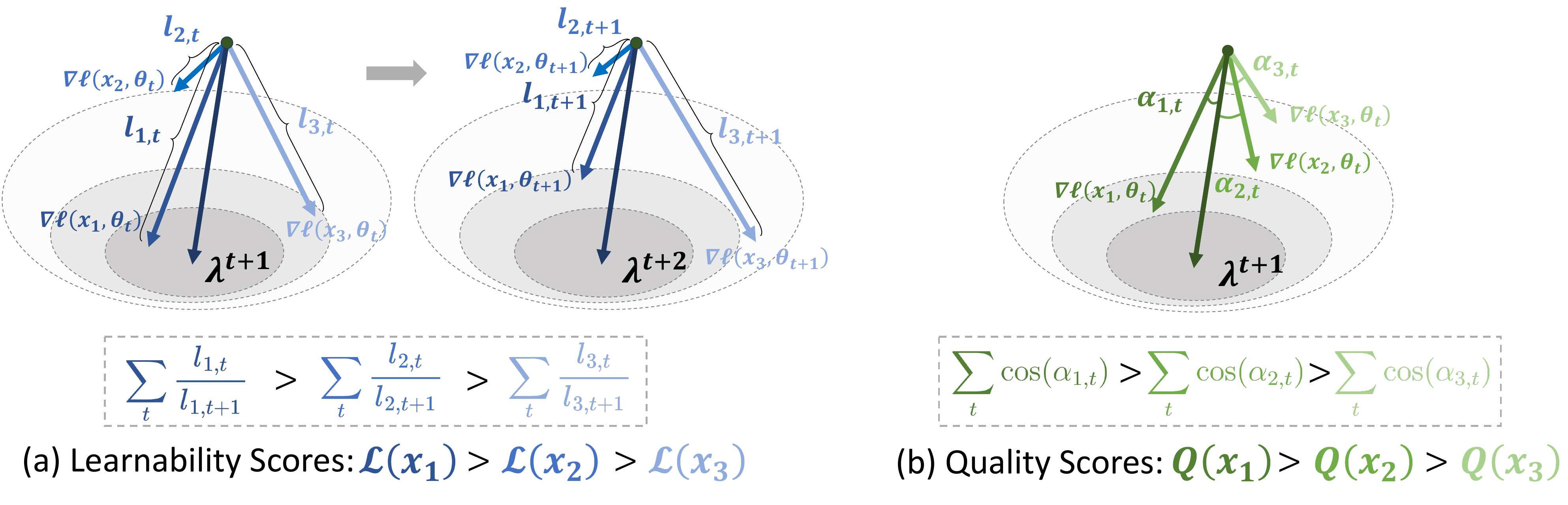}
\vspace{-0.6cm}
  \caption{
  Illustration of LQS scoring method. The left part demonstrates the calculation of the learnability score, and the right part depicts the computation of the quality score.
  }
  \label{fig:scoring}
\vspace{-0.3cm}
\end{figure}

The \textbf{\textit{learnability}} of each data sample represents the difficulty change during model training, as illustrated in Figure \ref{fig:scoring}a.
For training step $t$ from 1 to $T$, the learnability of a sample $\xtrn_n$ is defined as its ability to reduce the loss over time during training. 
The learnability is represented as:
\begin{equation}
\label{eq:learnability}
\mathcal{L}(\xtrn_n) = \sum_{t=1}^{T-1} \frac{l_{n,t}}{l_{n,t+1}} = \sum_{t=1}^{T-1} \frac{\|\nabla \ell(\xtrn_n, \vtd)\|}{\|\nabla \ell(\xtrn_n, \vtdp)\|},
\end{equation}
where $\nabla \ell(\xtrn_n, \vtheta_t)$ denotes the gradient of loss function for sample $\xtrn_n$ at training step $t$, with model parameters $\vtheta_t$, and $l_{n,t}$ is its magnitude.
A high learnability score indicates that the sample significantly reduces the loss in training, particularly if its gradient magnitude is initially high and decreases substantially over time. 
Such samples are challenging yet beneficial for training, making them more suitable for later stages of training.
Conversely, noisy samples or those with unstable gradients yield a low learnability score, enabling their identification and efficient filtering during data selection.

The \textbf{\textit{quality}} of each data sample contributes to data efficacy during model training, as depicted in Figure \ref{fig:scoring}b.
It is measured by the consistency of $\nabla \ell(\xtrn_n, \vtheta_t)$ with a target vector $\vldp$ in Equation \ref{eq:target vector}, which represents the average gradient of the loss function for all data at training step $t+1$. 
The quality score is computed as:
\begin{equation}
\text{Q}(\xtrn_n) = \sum_{t=1}^{T-1} \cos(\alpha_{n, t}) = \sum_{t=1}^{T-1} \frac{{\vldp}^\top \nabla \ell(\xtrn_n, \vtd)}{\|\vldp\| \cdot \|\nabla \ell(\xtrn_n, \vtd)\|},
\label{eq:quality}
\end{equation}
where $\alpha_{n, t}$ is the angle between two vectors.
A higher cosine similarity $\cos(\alpha_{n, t})$ indicates that the gradient convergence direction on $\xtrn_n$ is more aligned with the target objective $\vldp$, implying a stronger contribution to reducing the loss $J(\vtheta)$.
As defined in \cite{gu2024data}, the target vector $\vld$ is:
\begin{equation}
\vld = 
\begin{cases} 
\vldp + \nabla \objt - \eta\cdot\nabla^2 L(\vtd, \vga)\cdot\vlam_{t+1}, & \text{if } t < T \\
\nabla J(\vtheta_{t}), & \text{if } t = T 
\end{cases}
\label{eq:target vector}
\end{equation}
Finally, we combine \textit{learnability} and \textit{quality} into a unified function to score data samples.
For a detailed explanation of the formula, please refer to the Appendix.
The score vector $\vga$ is defined as:
\begin{equation}
\label{eq:score vector}
\vga = \left\{\gn \middle| \gn = \sum_{t=1}^{T-1}\frac{{\vldp}^\top\nabla \ell(\xtrn_n, \vtd)}{\|\nabla \ell(\xtrn_n, \vtdp)\|}, 1 \leq n \leq |\Dtrn| \right\}
\end{equation}

Larger $\gamma_n$ values indicate samples with higher quality and significant contributions to reducing the downstream loss $J(\vtheta)$, especially when introduced during later training stages.
In contrast, lower $\gamma_n$ values correspond to samples that are easy and less informative, better suited for early-stage training, or potentially noisy, which can be filtered out in data selection settings.
For detailed implementation of LQS, please refer to the Appendix.

\subsection{Data Ordering}
\label{sec: data ordering}

\subsubsection{Baseline Methods}
Existing methods usually utilize the shuffling method to organize training data.
Additionally, some approaches apply curriculum learning \citep{campos2021curriculum, kim2024strategic}, which can be considered a type of sorting method.

\textbf{Shuffling} method randomly arranges the order of training data to prevent inherent imbalanced distribution.
It acts as a baseline for data organization but does not account for data efficacy.

\textbf{Sorting} method sorts training data based on specific criteria, such as quality or difficulty.
The permutation function $\pi_{sort}$ provides the ranking index of each element in the training data in ascending order.
Curriculum learning is an example, mimicking the human learning process by starting with simpler samples and gradually increasing difficulty.
Although sorting-based approaches can enhance training efficacy, they may encounter issues like model forgetting, data distribution bias, and even data duplication, which can hinder performance.

\subsubsection{Our Method}
\begin{figure}[!hb]
\vspace{-0.5cm}
  \centering
  \includegraphics[width=1.0\columnwidth]{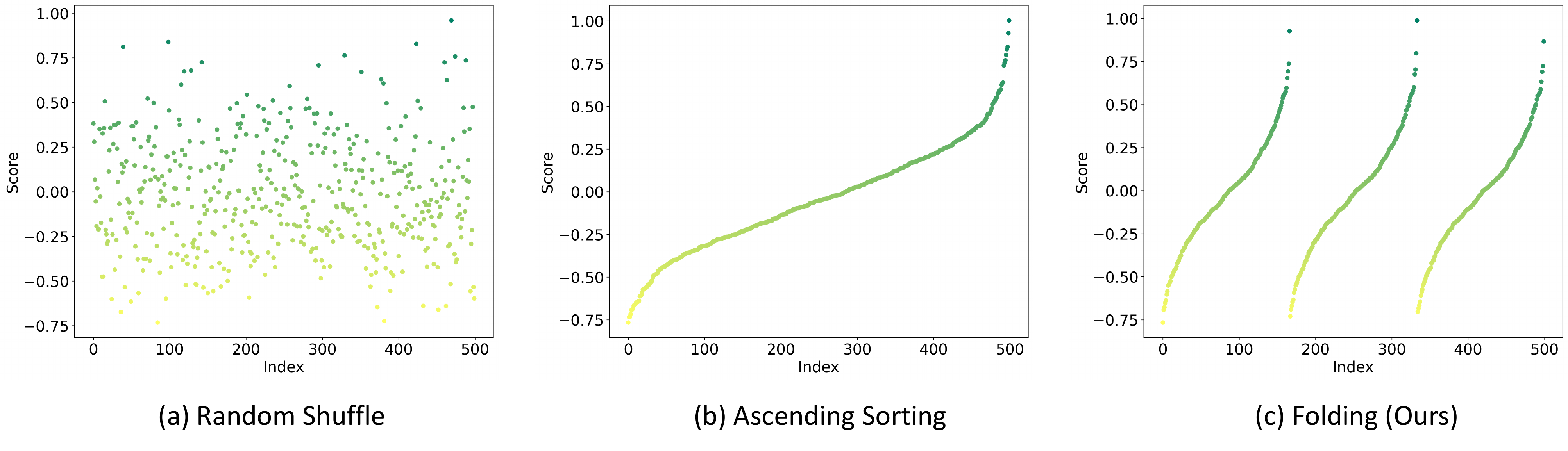}
\vspace{-0.5cm}
  \caption{
  Illustration of ordering methods.
  The right one is Folding method, an advanced multi-fold version of the Sorting method.
  These results are based on 500 random samples from RedPajama.
  }
  \label{fig:ordering}
\vspace{-0.8cm}
\end{figure}

\textbf{Folding} method is proposed to improve training data efficacy and address the negative influences brought by the sorting method.
The new method, named \textbf{\textit{folding learning}}, reorganizes the dataset by repeating the curriculum learning multiple times without duplication.
The repeated times is defined as the folding layers $L$.
As demonstrated in Figure \ref{fig:ordering}, the folding method samples sorted data $L$ times without replacement at a fixed interval $L$.
The permutation function $\pi_{fold}$ is defined in Equation \ref{eq:pi_fold}, while $\pi_{sort}$ is described in Equation \ref{eq:f_o}.
Folding learning not only inherits the benefits from curriculum learning but also mitigates issues of model forgetting, data distribution bias, and even duplication.
\begin{equation}
\pi_{fold}(\vga; L) = \bigcup_{\ell=0}^{L-1} \left\langle \pi_{sort}(\vga)_{i} \ \middle| \ i \in \{j \ |\ j \equiv \ell \pmod{L}, \, 1 \leq j \leq |\Dtrn| \} \right\rangle
\label{eq:pi_fold}
\end{equation}

\section{Experiment}

\subsection{Experimental Setup}
\label{sec:exp setting}
\textbf{Data.}
(1) \textbf{General data.}
We utilize the Redpajama \citep{weber2024redpajama} sourced from CommonCrawl as $\mathcal{D}$, which offers a relatively balanced knowledge distribution \citep{doremi}.
The downstream loss $J(\mathbf{\theta})$ for the data scoring model is computed on the LIMA \citep{lima}, which is a high-quality dataset with 1,030 diverse instruction-response pairs spanning various downstream scenarios.
(2) \textbf{Math data.}
We use the OpenWebMath \citep{paster2023openwebmath} as $\mathcal{D}$. 
The downstream loss $J(\mathbf{\theta})$ is computed on the MiniF2F~\citep{zheng2021minif2f}, which is a high-quality dataset consisting of 488 manually formalized mathematical problem statements, spanning multiple domains. 
(3) \textbf{Code data.} 
We employ The-Stack-v2 dataset \citep{lozhkov2024starcoder} as $\mathcal{D}$. 
The downstream loss $J(\mathbf{\theta})$ is computed on the Epicoder-380k \cite{wang2025epicoder}, which is a synthetic dataset with 380k diverse instruction-response pairs spanning multiple code generation scenarios. 

\textbf{Model.} 
We apply the Mistral~\citep{mistral} architecture for pre-training on general data, and the Qwen1.5~\citep{bai2023qwen} for post-training in the math and code domain respectively using the official pre-trained weights. 


\textbf{Training.}
Unless otherwise specified,
we pre-train all LMs for one epoch, using a batch size of 512 and a maximum input length of 1,024.
For the model, we utilize 160M parameters by default.
For pre-training, we randomly select a 1B-token subset from Redpajama by default, while for post-training, we sample a 1B-token subset each from OpenWebMath and The-Stack-v2.
This setting allows us to evaluate the impact of different data scoring and data ordering methods on LM training using $\mathcal{D}$.



\textbf{Baselines.} 
Based on the \textbf{DELT} pipeline, we compare the proposed methods with existing baselines for data scoring and ordering. 
As described in Section \ref{sec:method}, the methods include: 
1) Data scoring: KenLM \citep{heafield-2011-kenlm}, PDS \citep{gu2024data}; 
2) Data ordering: Shuffling (Random), Sorting (Curriculum Learning) \citep{campos2021curriculum}.

For additional details of the experimental setup, such as evaluation, please see the Appendix.

\subsection{Main Results}
\label{sec:main}
\textbf{Data Efficacy with Different Model Sizes and Data Scales.}
Table \ref{tab: model data size} presents the evaluation results of the pre-trained LMs on the OLMo \citep{olmo} evaluation benchmarks. 
As shown, DELT consistently outperforms the baselines on most datasets, achieving the best overall performance across models with 160M, 470M, and 1B parameters (Table \ref{tab: model size}a), as well as across data sizes of 1B, 10B, and 50B tokens (Table \ref{tab: data size}b). 
These results demonstrate that DELT steadily improves the data efficacy in LM training across various model sizes, data scales, and downstream tasks.
\input{Tabs/exp_model_size}

\textbf{Adaptability of Different Methods in DELT.}
Table \ref{tab:efficiencyvsefficacy} presents the results of different methods applied within the DELT paradigm, all of which significantly outperform the conventional baseline. 
Notably, regardless of whether data selection is applied, our proposed LQS scoring method achieves the best results. 
Furthermore, our proposed Folding ordering method consistently provides noticeable improvements across all baseline methods.
\input{Tabs/exp_efficiency_vs_efficacy}

\textbf{Data Efficiency Promotion on Existing Methods.}
Figure \ref{fig:efficiency_setting} further evaluates the data efficiency of the DELT framework by involving the data selection setting. 
Compared to previous methods that only consider data selection (Shuffling), the DELT framework (Sorting and Folding) achieves superior performance across the majority of selection ratios. 
The results show that the DELT framework is compatible with the data selection method, and the combination further improves their data efficiency.
\input{Tabs/exp_fig_pds_kenlm_ratio}

\textbf{Domain Robustness in Post-training (Table \ref{tab:exp_domain}).}
To validate the robustness of the DELT paradigm, we conduct post-training experiments on datasets of OpenWebMath \citep{paster2023openwebmath} and The-Stack-v2 \citep{lozhkov2024starcoder} respectively across math and code domains, both of which are sourced from the web data.
As shown in Table \ref{tab:exp_domain}, our method consistently outperforms the baselines across benchmarks in different domains, demonstrating the strong versatility of the DELT under our proposed LQS and FO methods.
\input{Tabs/exp_domain_generalization}




\textbf{Stability on Different Epochs (Figure \ref{fig:ablation eps}).}
In addition to one-epoch training, we also evaluate the effectiveness of DELT under a multi-epoch setting. 
As shown in Figure \ref{fig:ablation eps}, our method consistently boosts the results of conventional random ordering as the number of epochs increases. 
While conventional random ordering exhibits fluctuations and slow improvements after the second epoch, our approach demonstrates steady progress, showcasing the effectiveness and stability of our method in maintaining superior performance over multiple epochs.

\subsection{Ablation Study}
\textbf{Ordering Method (Table \ref{tab:ablation order}).}
Data ordering is a key component of the DELT framework and plays a significant role in improving data efficacy.
To verify the effectiveness of different ordering methods and identify the optimal one for DELT, we conduct corresponding experiments.
The results in Table \ref{tab:ablation order} show that ascending sorting improves the result while descending sorting leads to a decline.
Besides, the proposed Folding method shows the most improvement among all the ordering methods.
This indicates the rationality and importance of data ordering within the DELT framework.
\input{Tabs/exp_order}

\textbf{$L$ in Folding Learning (Figure \ref{fig:ablation layers}).}
We explore the influence of different folding layers $L$ on model performance in Figure \ref{fig:ablation layers}.
As $L > 1$, the performance consistently surpasses that of $L=1$ (curriculum learning), verifying the benefits brought by folding learning. 
The average performance initially increases and then gradually declines, peaking at $L=3$.
In the experiments conducted in this paper, $L$ is set to a default value of 3.
For more experimental details, please refer to the Appendix.
\input{Tabs/exp_fig_eps_layers}

\section{Conclusion}

Regarding the underexplored research on data efficacy, we propose a general paradigm, DELT, for enhancing data efficacy in language model training.
Meanwhile, we explore the relationship between data efficacy and efficiency, noting that their purposes differ significantly while they are related.
Our comprehensive experiments with various DELT implementations confirm the effectiveness of DELT, and our newly designed methods, respectively for data scoring and ordering, outperform the other methods.
We believe that the proposed paradigm highlights the potential of data efficacy as a field beyond existing methods, such as curriculum learning.
Moreover, extensive experiments reveal that data ordering and data selection are compatible, and their combined use can further boost performance.
This suggests the possibility of unifying the paradigms by incorporating data scoring, data selection, and data ordering.
By adopting this unified paradigm, it becomes feasible to simultaneously consider data efficacy and efficiency, paving the way for promising future directions.

\clearpage
{
    \small
    \bibliographystyle{unsrt}
    \bibliography{neurips_2025}

}


\newpage
\input{appendix}


\end{document}

%% file: math_commands.tex
\usepackage{amsmath,amsfonts,bm}
\usepackage{amsthm}
\usepackage{bbm}

\def\eqref#1{equation~(\ref{#1})}
\def\Eqref#1{Eq.~(\ref{#1})}

\def\1{\bm{1}}

\def\vtheta{{\bm{\theta}}}
\def\vgamma{{\bm{\gamma}}}
\def\vlambda{{\bm{\lambda}}}

\def\vh{{\bm{h}}}

\def\vw{{\bm{w}}}

\DeclareMathAlphabet{\mathsfit}{\encodingdefault}{\sfdefault}{m}{sl}
\SetMathAlphabet{\mathsfit}{bold}{\encodingdefault}{\sfdefault}{bx}{n}

\def\sR{{\mathbb{R}}}

\newcommand{\R}{\mathbb{R}}

\newcommand{\vtd}{\vtheta_t}
\newcommand{\vtdp}{\vtheta_{t+1}}

\newcommand{\prx}{\mathrm{prx}}
\newcommand{\sub}{\mathrm{sub}}
\newcommand{\gn}{\gamma_n}

\newcommand{\vga}{\vgamma}

\newcommand{\vgao}{\vga^*}

\newcommand{\vlam}{\vlambda}

\newcommand{\vld}{\vlam_t}
\newcommand{\vldp}{\vlam_{t+1}}

\newcommand{\vldo}{\vlam^*_t}
\newcommand{\vldop}{\vlam^*_{t+1}}

\newcommand{\Dtrn}{\mathcal{D}}

\newcommand{\Dprx}{\mathcal{D}^\prx}

\newcommand{\Dorder}{\mathcal{D}'}

\newcommand{\gan}{\gamma_{n}}

\newcommand{\xtrn}{x}

\newcommand{\objt}{J(\vtheta_t)}
\newcommand{\Dselect}{\mathcal{D}^\sub}

%% file: Tabs/exp_model_size.tex
\begin{table}[htbp!]
\vspace{-0.5cm}
\centering
\scriptsize

\caption{
Efficacy results on different downstream benchmarks.
The conventional method presents the average result over three random seeds in this and the following tables.
Ours means applying LQS for data scoring and Folding for data ordering within the DELT paradigm.
}
\vspace{-0.8cm}
\begin{subtable}
\centering
\caption*{(a) Results (\%) for 1B-token data across model sizes (160M, 470M, 1B).}
\label{tab: model size}
\begin{tabular}{c|cccccccc|c}
\toprule
\multicolumn{1}{l}{} & ARC-c            & ARC-e            & HS               & LAMB             & OBQA             & PIQA             & SciQ             & Wino             & Avg.             \\
\midrule
\multicolumn{1}{l}{} & \multicolumn{9}{c}{Model size = 160M}                            \\ \midrule
Conventional         & 21.27          & 34.32          & 27.85          & 20.25          & 24.40          & 55.19 & 56.93          & 50.72          & 36.37          \\
Ours                 & \textbf{21.59}          & \textbf{36.07} & \textbf{28.41} & \textbf{23.79} & \textbf{25.60} & \textbf{56.37}          & \textbf{59.80} & \textbf{53.04} & \textbf{38.08} \\
\midrule
\multicolumn{1}{l}{} & \multicolumn{9}{c}{Model size = 470M}                            \\ \midrule
Conventional         & 21.16         & 34.91          & 28.11          & 21.88          & 23.90          & 56.07          & 58.75          & 50.04          & 36.85          \\
Ours                 & \textbf{22.33} & \textbf{35.88} & \textbf{28.45} & \textbf{23.26} & \textbf{26.60} & \textbf{57.20} & \textbf{60.10} & \textbf{52.81} & \textbf{38.33}       \\
\midrule
\multicolumn{1}{l}{} & \multicolumn{9}{c}{Model size = 1B}         
\\ \midrule
Conventional    &20.58   & 36.12   & 28.32      & 23.56     & 25.00     & 56.49     & 60.05     & \textbf{52.07}     & 37.77 \\
Ours                 & \textbf{22.76} & \textbf{37.95} & \textbf{29.95} & \textbf{26.38} & \textbf{26.00} & \textbf{58.07} & \textbf{60.90} & 51.28 & \textbf{39.17} \\
\bottomrule
\end{tabular}
\end{subtable}
\begin{subtable}
\centering
\caption*{(b) Results (\%) for 160M model across data sizes (10B, 50B).}
\label{tab: data size}
\begin{tabular}{c|cccccccc|c}
\toprule
\multicolumn{1}{l}{} & ARC-c            & ARC-e            & HS               & LAMB             & OBQA             & PIQA             & SciQ             & Wino             & Avg.             \\
\midrule
\multicolumn{1}{l}{} & \multicolumn{9}{c}{Data size = 10B tokens}                                                                                                                               \\
\midrule
Conventional         & 22.82          & 38.51          & 30.72          & 30.40          & 25.70          & 57.32          & 64.90          & 51.54          & 40.24          \\
Ours                 & \textbf{24.38} & \textbf{39.80} & \textbf{31.64} & \textbf{32.98} & \textbf{27.21} & \textbf{58.56} & \textbf{66.70} & \textbf{51.67} & \textbf{41.62} \\
\midrule
\multicolumn{1}{l}{} & \multicolumn{9}{c}{Data size = 50B tokens}                                                                                                                               \\
\midrule
Conventional         & 24.06 & \textbf{41.88} & 32.05 & 33.79 & 26.80 & 58.11 & \textbf{69.00} & 51.93 & 42.20        \\
Ours                 & \textbf{24.65} & 41.07 & \textbf{33.00}    & \textbf{36.07} & \textbf{29.30} & \textbf{59.10}  & 68.40 & \textbf{52.67} & \textbf{43.03}         \\
\bottomrule
\end{tabular}
\end{subtable}
\label{tab: model data size}
\vspace{-0.7cm}
\end{table}

%% file: Tabs/exp_efficiency_vs_efficacy.tex
\begin{table}[hbtp!]
\vspace{-0.3cm}
\centering
\scriptsize
\setlength{\tabcolsep}{5pt}
\caption{
Efficacy results of different DELT implementations.
The best scores for each model size are highlighted in \textbf{bold}, while the second-best scores are shown in \textbf{\textit{italic bold}}.
The selection methods report the highest scores across all selection ratios.
}
\begin{tabular}{p{1.2cm}|p{1.0cm}p{0.7cm}p{0.7cm}|cccccccc|c}
\toprule
Pipeline                         & Scoring & Selection & Ordering & ARC-c            & ARC-e            & HS               & LAMB             & OBQA             & PIQA             & SciQ             & Wino             & Avg.             \\
\midrule
Conventional                     & -       & -         & -        & 21.27          & 34.32          & 27.85          & 20.25          & 24.40          & 55.19          & 56.93          & 50.72          & 36.37          \\
\midrule
\multirow{8}{*}{DELT}            & KenLM   & -         & Sorting        & 21.93 & 33.96          & \textbf{\textit{28.09}}          & 20.69          & 25.20          & 54.79          & 56.20          & 50.59          & 36.43             \\
                                 & KenLM   & -         & Folding        & 20.98 & 35.00 & 28.02 & 22.55 & 23.90 & \textbf{\textit{56.54}} & 58.30 & \textbf{\textit{51.36}} & 37.08             \\
                                 & PDS     & -         & Sorting       & \textbf{\textit{22.44}}          & 34.18          & 27.98          & 21.35          & 25.40          & 55.28          & 55.80          & 49.17          & 36.45          \\
                                 & PDS     & -         & Folding  & 21.93          & 34.81          & 28.04          & 22.43          & \textbf{26.00} & 56.42           & \textbf{\textit{59.30}}          & 50.20          & 37.40          \\
                                 & LQS & -         & Sorting  & \textbf{23.22} & \textbf{\textit{35.24}} & 28.03 & \textbf{\textit{22.79}} & 24.70 & \textbf{56.85} & 57.90 & 51.17 & \textbf{\textit{37.49}} \\     
                                 & LQS & -         & Folding  & 21.59          & \textbf{36.07} & \textbf{28.41} & \textbf{23.79} & \textbf{\textit{25.60}}          & 56.37          & \textbf{59.80} & \textbf{53.04} & \textbf{38.08} \\     
                                 
                                 \cmidrule{2-13}
                                 
                                 & KenLM   & \checkmark       & Sorting        & 21.93            & 34.68            & 27.78            & 19.37            & \textbf{26.40}             & 54.95            & 56.30             & \textbf{\textit{52.96}}            & 36.80          \\
                                 & KenLM   & \checkmark       & Folding        & \textbf{\textit{22.10}} & 34.30 & 27.62 & 21.56 & 25.00 & 56.26 & 58.10 & 52.80 & 37.22          \\
                                 & PDS     & \checkmark       & Sorting       & \textbf{22.61} & 35.27          & \textbf{\textit{28.08}}          & 19.68          & \textbf{\textit{25.80}}         & \textbf{56.53} & 59.60          & 51.54          & 37.38          \\
                                 & PDS     & \checkmark       & Folding       & 21.66 & \textbf{\textit{36.01}} & 28.05 & \textbf{24.33} & 24.10 & 55.61 & \textbf{61.70} & 52.47 & \textbf{\textit{37.99}}          \\
                                 & LQS & \checkmark       & Sorting       & \textbf{\textit{22.10}}             & 35.61            & 28.05            & 22.53            & 23.60            & 55.93            & 59.60             & 51.38            & 37.35            \\
                                 & LQS & \checkmark       & Folding  & 21.59          & \textbf{36.07} & \textbf{28.41} & \textbf{\textit{23.79}} & 25.60          & \textbf{\textit{56.37}}          & \textbf{\textit{59.80}} & \textbf{53.04} & \textbf{38.08}           \\ 
\bottomrule
\end{tabular}
\label{tab:efficiencyvsefficacy}
\vspace{-0.2cm}
\end{table}

%% file: Tabs/exp_fig_pds_kenlm_ratio.tex
\begin{figure}[!hb]
\vspace{-0.2cm}
  \centering
  \includegraphics[width=0.925\columnwidth]{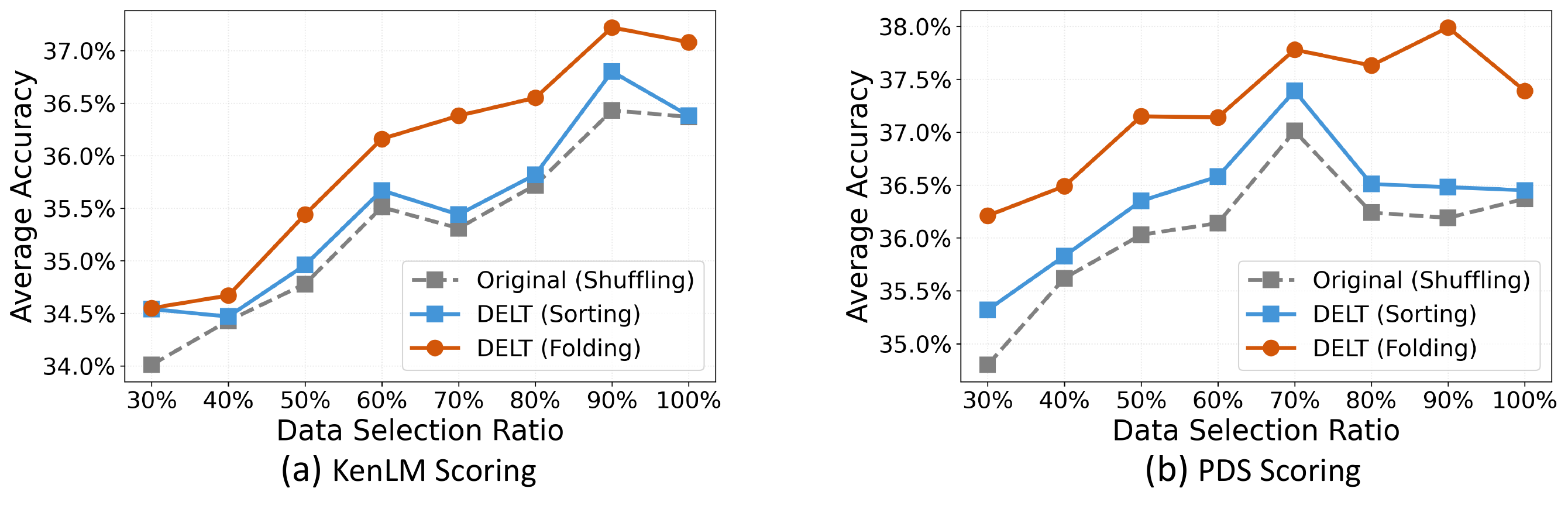}
\vspace{-0.3cm}
  \caption{
    The performance of KenLM \citep{heafield-2011-kenlm} and PDS \citep{gu2024data} under different data selection ratios, both with and without the DELT paradigm.
    Data efficiency is enhanced when integrated into DELT.
  }
  \label{fig:efficiency_setting}
\vspace{-0.3cm}
\end{figure}

%% file: Tabs/exp_domain_generalization.tex
\begin{table}[hbtp!]
\vspace{-0.3cm}
  \scriptsize
  \centering
  \caption{
  Efficacy of models trained on different domain-specific datasets.          
  }

  \begin{minipage}{0.48\textwidth} %
    \centering
    \label{tab:exp_code}
    \caption*{(a) Results on code domain.}
    \begin{tabular}{llll}
    \toprule
    \multicolumn{1}{l}{}      & Method       & \multicolumn{1}{l}{HumanEval} & \multicolumn{1}{l}{MBPP} \\
    \midrule
    \multirow{2}{*}{Qwen1.5-0.5B} & Conventional & 7.00                             & 7.93                      \\
                              & Ours         & \textbf{9.76}                             & \textbf{9.40}                       \\
                              \midrule
    \multirow{2}{*}{Qwen1.5-1.8B} & Conventional & 9.15                             & 12.00                      \\
                              & Ours         & \textbf{16.46}                            & \textbf{13.20}      \\
                              \bottomrule
    \end{tabular}
  \end{minipage}
  \hfill
  \begin{minipage}{0.48\textwidth}
    \centering
    \label{tab:exp_math}
    \caption*{(b) Results on math domain.} 
    \begin{tabular}{llll}
    \toprule
    \multicolumn{1}{l}{}      & Method       & \multicolumn{1}{l}{MathQA} & \multicolumn{1}{l}{GPQA Diamond} \\
    \midrule
    \multirow{2}{*}{Qwen1.5-0.5B} & Conventional & 21.23                            & 24.92                      \\
                              & Ours         & \textbf{22.73}                            & \textbf{26.83}                      \\
                              \midrule
    \multirow{2}{*}{Qwen1.5-1.8B} & Conventional & 22.72                            & 27.17                      \\
                              & Ours         & \textbf{24.75}                            & \textbf{28.94}      \\
                              \bottomrule
    \end{tabular}
  \end{minipage}
  
  \label{tab:exp_domain}
\vspace{-0.1cm}
\end{table}


%% file: Tabs/exp_order.tex
\begin{table}[htbp!]
\vspace{-0.2cm}
\centering
\scriptsize
\caption{Comparison among different ordering methods.
Sorting$_{asc}$ refers to an \textbf{ascending} sorting by scores (from low to high), while sorting$_{des}$ denotes a \textbf{descending} sorting (from high to low).
}
\label{tab:ablation order}
\begin{tabular}{c|c|cccccccc|c}
\toprule
\multicolumn{1}{l}{}  & Ordering           & ARC-c          & ARC-e          & HS             & LAMB           & OBQA       & PIQA           & SciQ          & Wino           & Avg.           \\
\midrule
Conventional & - & 21.27          & 34.32          & 27.85          & 20.25          & 24.40          & 55.19 & 56.93          & 50.72          & 36.37 \\
\midrule
\multirow{3}{*}{PDS}  &  Sorting$_{des}$         & 22.01          & 33.67          & 27.77          & 14.67          & 23.80       & 54.90           & 51.90          & \textbf{51.93}          & 35.08$\downarrow$          \\
                      & Sorting$_{asc}$          & \textbf{22.44}          & 34.18          & 27.98          & 21.35          & 25.40       & 55.28          & 55.80          & 49.17          & 36.45$\uparrow$          \\
                      & Folding                 & 21.93 & \textbf{34.81} & \textbf{28.04} & \textbf{22.43}          & \textbf{26.00}         & \textbf{56.42} & \textbf{59.30} & 50.20  & \textbf{37.39}$\uparrow$          \\
\midrule
\multirow{3}{*}{LQS} & Sorting$_{des}$      & 20.69 & 34.72 & 27.78 & 21.20 & 23.10 & 56.12 & 58.20 & 49.11 & 36.36$\downarrow$              \\
                      & Sorting$_{asc}$ & \textbf{22.18}          & 35.40           & 28.01          & 23.48  & 23.80       & 55.60           & 56.80          & 51.07          & 37.04$\uparrow$          \\
                      & Folding                 & 21.59          & \textbf{36.07} & \textbf{28.41} & \textbf{23.79} & \textbf{25.60}       & \textbf{56.37}          & \textbf{59.80} & \textbf{53.04} & \textbf{38.08}$\uparrow$ \\
\bottomrule
\end{tabular}
\vspace{-0.1cm}
\end{table}


%% file: Tabs/exp_fig_eps_layers.tex
\begin{figure*}[htbp]
\vspace{-0.1cm}
  \begin{minipage}[c]{0.45\linewidth}
    \centering
    \includegraphics[width=0.90\linewidth]{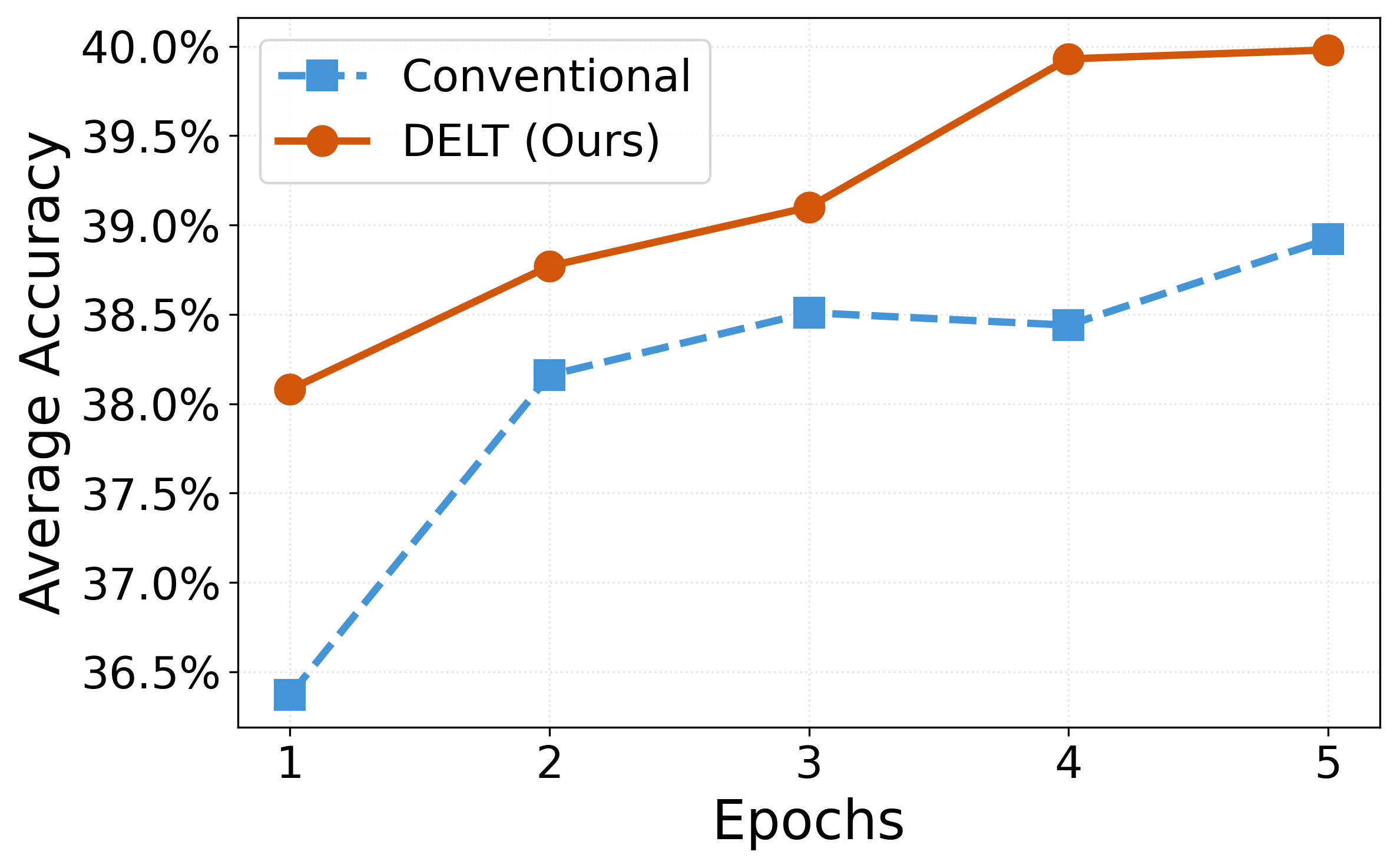}
    \caption{
    Performance on different epochs.
    Benchmarks from OLMo \citep{olmo} are applied.
    }
    \label{fig:ablation eps}
  \end{minipage}
  \hspace{1.0cm}
  \begin{minipage}[c]{0.45\linewidth}
    \centering
    \includegraphics[width=0.90\linewidth]{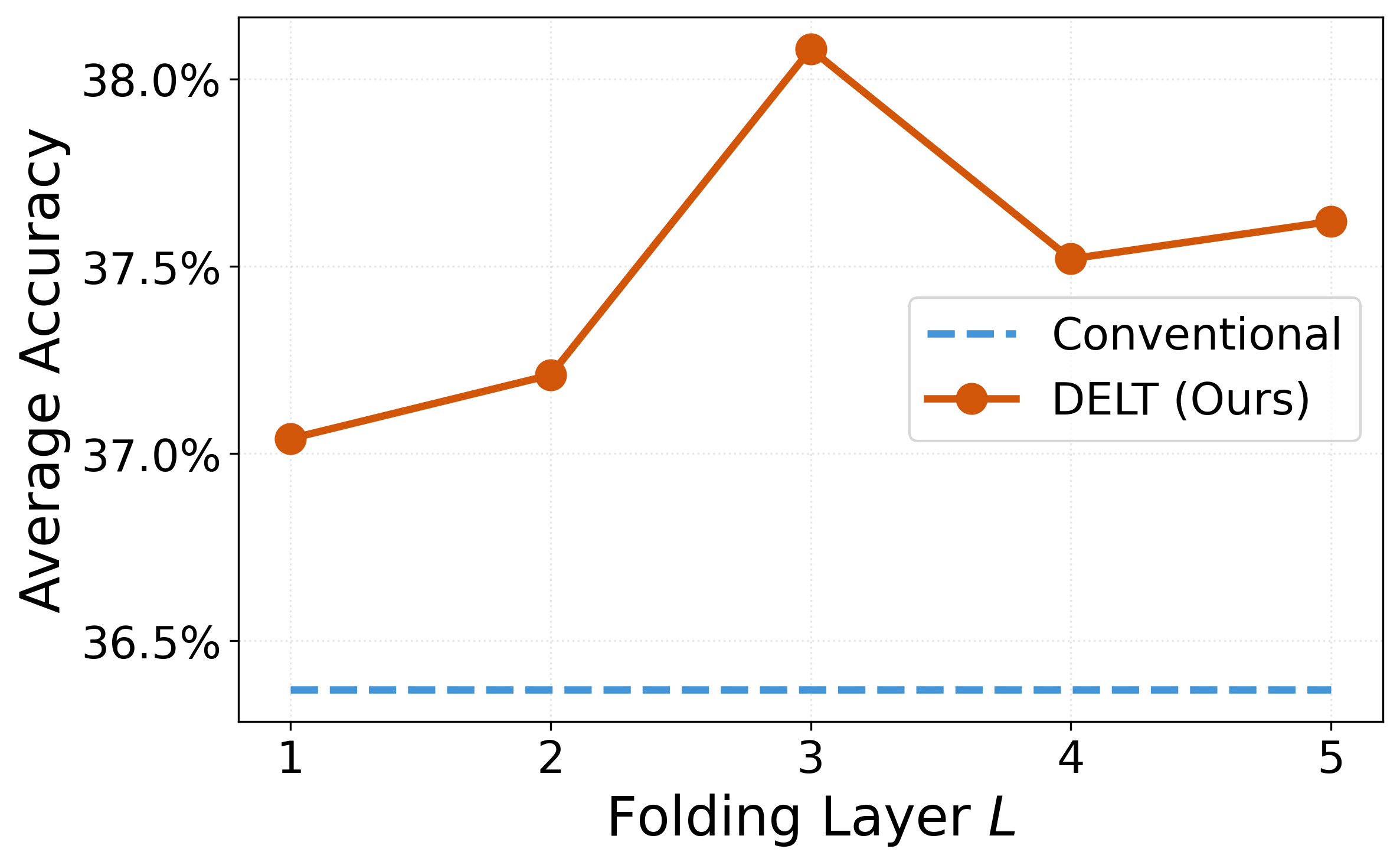}
    \caption{
    Influence of the folding layers $L$.
    Benchmarks from OLMo \citep{olmo} are applied.
    }
    \label{fig:ablation layers}
  \end{minipage}
\vspace{-0.2cm}
\end{figure*}

%% file: appendix.tex
\appendix
\begin{center}
    {\large \textbf{Appendix: Data Efficacy for Language Model Training}}
\end{center}
\section{Additional Experimental Setup}
\label{sec: ad exp setup}


\textbf{Evaluation.} 
For general evaluation, we assess the trained models on a range of standard natural language understanding and reasoning benchmarks, including Hellaswag (HS; \citealp{hella_swag}), Winogrande (Wino; \citealp{winogrande}), LAMBADA (LAMB; \citealp{lambada}), OpenbookQA (OBQA; \citealp{openbookqa}), ARC-easy/challenge (ARC-e/c;~\citealp{arc}), PIQA \citep{piqa}, SciQ \citep{sciq}, and BoolQ \citep{boolq}.
For domain-specific tasks, we assess the models on mathematical reasoning and code generation benchmarks. Specifically, for math, we use GPQA Diamond \citep{rein2024gpqa} and MathQA \citep{amini2019mathqa}, while for code, we use HumanEval \citep{chen2021humaneval} and MBPP \citep{austin2021mbpp}.

For the code generation benchmarks, we use a 0-shot setting for HumanEval and a 3-shot setting for MBPP. 
Code generation performance is reported using pass rate@1, which indicates the percentage of first-attempt solutions that pass all associated unit tests (\texttt{pass@1}). 
For the other benchmarks, the tasks are framed as multiple-choice questions, where the model selects the correct answer by minimizing the normalized loss across all candidate options (\texttt{acc\_norm}).

\textbf{Compute Resources.}
We train the 160M model on 1B-token datasets using a single NVIDIA A100 40GB GPU. For experiments with the 160M, 470M, and 1B models on 10B and 50B-token datasets, we utilize 8 NVIDIA A100 40GB GPUs. All data scoring steps, including proxy data annotation, scorer training, and data scoring, are performed on a single NVIDIA A100 80GB GPU.

\textbf{Training Configuration.}
To create the data scorer, we fine-tune the Fairseq-Dense-125M model \citep{fairseq_dense} on the solved data weights $\vga$. As described in algorithm \ref{alg:method}, we apply a linear transformation to the mean-pooled representations of instances along the sequence length. The hidden state size is set to 768. The optimization of algorithm \ref{alg:method} is performed using the AdamW optimizer \citep{adamw} for 5 epochs, with a learning rate of $1\times10^{-4}$ and a batch size of 512. 
We train on 90\% of the samples and reserve the remaining 10\% from $\Dprx$ as a validation set. The checkpoint with the highest Spearman correlation score \citep{de2016spearman,gu2024data} on the validation set is selected to infer data quality scores in $\Dtrn$.

For LMs training, all models are trained with a batch size of 256 and a maximum input sequence length of 1,024 for one epoch. The AdamW optimizer \citep{adamw} is paired with a cosine learning rate scheduler. 
The scheduler includes a warm-up phase for the first 2,000 steps, after which the learning rate decays to 10\% of its peak value.
The model architecture and corresponding learning rates are summarized in Table \ref{tab:model_config}, following the configurations in \citep{gu2024data}.

\input{Tabs/appendix_model_config}

\section{Societal Impact}
Our work introduces DELT, the first data efficacy paradigm that enhances language model performance through strategic data ordering.
By unifying data scoring, selection, and ordering, DELT improves both the efficacy and efficiency in LM training, including both pre-training and post-training.
This foundational advancement enables performance improvements without altering model size or dataset scale, offering substantial positive societal impacts for the development of general AI.

Meanwhile, DELT can reduce the computational resources and energy consumption required for training language models, contributing to sustainability in AI development. Furthermore, the enhanced efficiency increases accessibility, enabling researchers and practitioners with limited resources to leverage advanced language models. By fostering the development of more robust and reliable models, this work can significantly benefit various societal domains, such as education and healthcare.

\section{Limitations and Future Work}
The proposed paradigm has two primary limitations.
1) The current verification is specifically focused on language models, with no evaluation in other modalities like image and audio, which depend on different scoring implementations.
2) Similar to PDS data scoring, the implementation of our designed LQS method requires calculating the downstream loss 
$J(\mathbf{\theta})$ on a high quality and small scale dataset.
In the future, we plan to scale up our method on larger models (e.g., tens or hundreds of billions of parameters) and larger datasets (terabyte level). 
Additionally, 
we aim to explore simpler and more effective data scoring methods and extend this paradigm to multimodal models.

\section{LQS Explanation}
\label{sec:explain_equation}               
We provide a more detailed explanation of the Learnability-Quality Scoring (LQS).
As described in Section \ref{sec:lqs method}, the Learnability Score $\mathcal{L}(\xtrn_n)$ (Equation \ref{eq:learnability}) and the Quality Score $\text{Q}(\xtrn_n)$ (Equation \ref{eq:quality}) are proposed, both of which are directly related to the training sample $\xtrn_n$ and are used as metrics to evaluate $\xtrn_n$. 

However, the reliability of these scores is directly influenced by the capability of the model checkpoint. Models with stronger capabilities produce more reliable scores. This capability is linked to the target vector, which is calculated as an average across all samples. When the structural parameters of the model remain unchanged, a stronger model shows greater consistency in the gradient directions of all samples concerning the model parameters, resulting in a larger $\nabla J(\vtheta_{t})$ (Equation \ref{eq:target vector}).
This corresponds to a greater magnitude of the target vector.
Therefore, we use the magnitude of the target vector to directly measure the model's capability, referred to as the reliability score in Equation \ref{eq:rel}.

\begin{equation}
\label{eq:rel}
\text{R}(\vtdp) = \|\vldp\|
\end{equation}

Finally, we anticipate that a stronger model will assign more weight to the scores. LQS is expressed as a combination of the sample's learnability-quality score and the model's capabilities.
\begin{align}
\gn &= \text{R}(\vtdp) \cdot \text{Q}(\xtrn_n) \cdot \mathcal{L}(\xtrn_n) \\ 
&= \sum_{t=1}^{T-1} \|\vldp\| \cdot
\frac{{\vldp}^\top \nabla \ell(\xtrn_n, \vtd)}{\|\vldp\| \cdot \|\nabla \ell(\xtrn_n, \vtd)\|} 
\cdot \frac{\|\nabla \ell(\xtrn_n, \vtd)\|}{\|\nabla \ell(\xtrn_n, \vtdp)\|}  \nonumber \\
&= \sum_{t=1}^{T-1} \|\vldp\| \cdot \frac{{\vldp}^\top \nabla \ell(\xtrn_n, \vtd)}{\|\vldp\| \cdot \|\nabla \ell(\xtrn_n, \vtdp)\|} \\
&= \sum_{t=1}^{T-1} \frac{{\vldp}^\top \nabla \ell(\xtrn_n, \vtd)}{\|\nabla \ell(\xtrn_n, \vtdp)\|}
\label{eq:lqs_ap}
\end{align}

\section{LQS Implementation}
\label{sec: score imp}

Similar to \citep{gu2024data}, our data scoring method follows these steps:
\textbf{1) Proxy data sampling}. A proxy dataset $\Dprx$ is first uniformly sampled from the pre-training corpus $\Dtrn$, serving as a representative subset of the larger corpus.
\textbf{2) Proxy data annotation.} 
We apply Eq. \ref{eq:score vector} to compute data scores for each instance in $\Dprx$, obtaining a set of data samples with scores as ground truth (see Section \ref{sec: score gt}).
\textbf{3) Data scorer training.} 
The data scorer, typically a small LM, is fine-tuned on the automatically annotated data samples in $\Dprx$ to predict data scores effectively.
(see Section \ref{sec: data scorer training})
\textbf{4) Full data scoring.} 
The trained data scorer is then applied to infer scores for the entire pre-training corpus $\Dtrn$.

\subsection{Proxy Data Annotation}
\label{sec: score gt}
To construct the ground truth scores $\vgao$ for $\Dprx$, we are inspired by \citep{gu2024data} and adopt a bi-level optimization framework (see Algorithm \ref{alg:method}) that quantifies the contribution of each data point in $\Dprx$ to the downstream performance. 
The goal is to determine an optimal score vector $\vgao = [\gamma^{*}_1, \gamma^{*}_2, \cdots, \gamma^{*}_{|\Dtrn|}]^\top$.

The bi-level optimization consists of two nested loops: 1) a forward loop that simulates the model training process, and 2) a reverse loop that adjusts the scores $\vga$ based on the model parameters at each training step. 
Specifically, in the \textbf{forward loop}, the model is trained for $T$ steps using gradient descent, where the training loss is weighted by the current scores $\vgao$. This process updates the model parameters $\vtheta_t$ iteratively, producing a trajectory of checkpoints $\vtheta_t$ from $t = 0$ to $t = T-1$.
In the \textbf{reverse loop}, the target vector $\vldo$ is computed through the training trajectory from $t = T-1$ to $t = 0$ according to Eq. \ref{eq:target vector}. 
where $\vldo$ represents the backward-propagated gradient at step $t$, $\nabla l(\xtrn_n, \vtheta_t)$ is the gradient of the loss for the $n$-th data point.
After each update, the scores are projected onto the probability simplex to ensure they remain valid probabilities $\operatorname{Proj}[\vgao]$.


\begin{algorithm}[H]
\small
\caption{Proxy Data Annotation}\label{alg:method}
\begin{algorithmic}
    \Require LM learning rate $\eta$.  Proxy data $\Dprx$. 
    Downstream loss $J(\vtheta)$. 
    Training steps $T$. $\operatorname{Proj}[\cdot]$ that projects a point in $\sR^{|\Dtrn|}$ to $U$. Model initialization $\vtheta_0$. 
    \Ensure Data quality scores $\vgao$.
    \State $\vgao = \left[\gamma^{*}_{1}, \gamma^{*}_{2}, \cdots, \gamma^{*}_{|\Dtrn|}\right] \gets \left[\frac{1}{|\Dprx|}, \frac{1}{|\Dprx|}, \cdots, \frac{1}{|\Dprx|}\right]$;
    \For{$t=0,1,\cdots, T-1$} \Comment{Forward loop}
    \State $\vtheta_{t+1} \gets \vtd - \eta\nabla L(\vtd, \vga)$ 
    \EndFor
    \State $\vlam_T \gets \nabla J(\vtheta_{T})$
    \For{$t=T-1,T-2,\cdots,1$} \Comment{Reverse loop}
    \State $\vldo \gets \vldop + \nabla \objt - \eta\nabla^2 L(\vtd, \vgao)\vldop$ 
    \Comment{\Eqref{eq:target vector}}
    \EndFor
    \For{$n=1,2,\cdots,|\Dtrn|$}
    \State $\gan^* \gets \gan^* + \alpha \sum_{t=1}^{T-1} \frac{{\vldop}^{\top} \nabla l(\xtrn_n, \vtd)}{\|\nabla l(\xtrn_n, \vtdp)\|}$ 
    \Comment{\Eqref{eq:score vector}}
    \EndFor
    \State $\vgao \gets \operatorname{Proj}\left[\vgao\right]$
    \State \Return $\vgao$
\end{algorithmic}
\end{algorithm}
\vspace{-0.7cm}

\subsection{Data Scorer Training}
\label{sec: data scorer training}
After we obtain the scores $\vgao$ for $\Dprx$, we then train a small LM, initialized from a pre-trained checkpoint, with a linear head to serve as the data scorer. 
The scorer is optimized on the proxy dataset $\Dprx$ to fit the ground-truth scores $\vga$ from Section \ref{sec: score gt}.
Specifically, each instance $x_n^\prx \in \Dprx$ is encoded by averaging the LM's output hidden states along the sequence, producing a feature representation $\overline{\vh}(x_n^\prx, \bm{\phi}) \in \R^d$, where $\bm{\phi}$ are the LM parameters and $d$ is the hidden state size. 
This representation is passed through a linear head, with parameters $\vw \in \R^d$ and $b \in \R$, to produce a predicted score. 
The parameters of the LM and linear head are optimized together using the Mean Squared Error (MSE) loss:
\begin{equation}
    \mathcal{L}_{\text{MSE}} = \frac{1}{|\Dprx|} \sum_{n=1}^{|\Dprx|} \left( \vw^\top \overline{\vh}(x_n^\prx, \bm{\phi}) + b - \gamma^*_n \right)^2,
\end{equation}
The optimal parameters $\bm{\phi}^*$, $\vw^*$, and $b^*$ are obtained by minimizing this loss. Once trained, the data scorer predicts scores for $\xtrn_n \in \Dtrn$ as
$\gamma(\xtrn_n) = {\vw^*}^\top \overline{\vh}(\xtrn_n, \bm{\phi^*}) + b^*$.
This process enables the data scorer to generalize from $\Dprx$ to the larger pre-training corpus $\Dtrn$ effectively.

\section{Samples Scored by LQS}
\label{sec:samples}
To better illustrate our advantages, we visualized part samples.  
Specifically, we randomly sampled examples from the top 10\% and bottom 10\% based on their scores for visualization.

As shown in Table \ref{tab:vis top10 eg12 ap} and \ref{tab:vis top10 eg34 ap}, high-scoring samples are high-quality, complex sentences that are not only challenging but also highly learnable, significantly aiding the model’s optimization in later stages.  
In contrast, as shown in Table \ref{tab:vis bottom10 ap}, low-scoring samples are often noisy and provide little to no benefit for model training, making them ideal candidates for filtering (see Table \ref{tab:vis bottom10 ap} example 1 and 2).  
However, some low-scoring samples consist of simple words or short phrases (see Table \ref{tab:vis bottom10 ap} example 3 and 4), which are beneficial for the model’s learning in the early stages.
\input{Tabs/sample_examples}

\section{More Experiments Results}
\textbf{Data Efficacy across Different Model Sizes and Data Scales on More Methods. (Table \ref{tab: appendix model data size}).}
To supplement Table \ref{tab: model data size}, we provide additional experimental results comparing DELT with data selection methods across different model sizes and data scales.  
As shown in Table \ref{tab: appendix model data size}, with increasing model parameters and data scales, our proposed DELT demonstrates consistent improvements and outperforms various methods across most benchmarks.
\input{Tabs/exp_tab1_ap}

\textbf{Comparison of Data Efficiency (Table \ref{tab:efficiencyvsefficacy_ap}).} 
To supplement Table \ref{tab:efficiencyvsefficacy}, we further highlight the superiority of the DELT framework by comparing it with the data efficiency setting.  
Table \ref{tab:efficiencyvsefficacy_ap} presents results for data processed through the efficiency (data selection) and efficacy (DELT) pipelines. 
Compared to the results from the conventional pipeline and the best-performing data efficiency pipeline (Selection \checkmark, Ordering -), the DELT framework, which incorporates both data selection and ordering (Selection \checkmark, Ordering \checkmark), consistently demonstrates significant improvements across various method combinations.
\input{Tabs/exp_tab2_ap}

\textbf{Stability on Different Epochs. (Table \ref{tab:ep_ap})}
To supplement Figure \ref{fig:ablation eps}, we further report the detailed results of the proposed DELT across different epochs on various benchmarks.  
As shown in Table \ref{tab:ep_ap}, with an increasing number of epochs, our method demonstrates stable improvements across most benchmarks, further highlighting its robustness and generalizability.
\input{Tabs/exp_epochs_ap}

\textbf{Details for $L$ in Folding Learning (Table \ref{tab:ablation l ap}).} 
To supplement Figure \ref{fig:ablation layers}, we provide detailed results for the proposed Folding Learning method with varying values of the parameter $L$.  
As shown in Table \ref{tab:ablation l ap}, model performance reaches its peak at $L=3$ and demonstrates significant advantages across most benchmarks.  
Notably, compared to traditional Curriculum Learning ($L=1$), our proposed method ($L>1$) achieves substantially better performance on all benchmarks.

\input{Tabs/exp_ablation_l}



%% file: Tabs/appendix_model_config.tex
\begin{table}[h]
    \centering
    \begin{tabular}{l|cccccc}
    \toprule
    Model Size     & $d_{\text{model}}$ & $d_{\text{FFN}}$ & $n_{\text{layers}}$ & $n_{\text{head}}$ & $d_{\text{head}}$ & learning rate \\
     \midrule
    160M           &  768               &   3,072          & 12                  &  12               &  64               &  $6\times 10^{-4}$               \\ 
    470M           &  1,024             &   4,096          & 24                  &  16               &  64               &  $3\times 10^{-4}$               \\ 
    1B             &  1,536             &   6,144          & 24                  &  16               &  96               &  $2.5\times 10^{-4}$               \\ 
    1.7B           &  2,048             &   8,192          & 24                  &  16               &  128              &  $2\times 10^{-4}$               \\ 
     \bottomrule
    \end{tabular}
    \vspace{3mm}
    \caption{Model configurations and corresponding learning rates.}
    \label{tab:model_config}
\end{table}
\vspace{-1cm}

%% file: Tabs/sample_examples.tex
\begin{tcolorbox}[title= Examples of Randomly Sampled High-Scoring Data Points (Top 10\%)]
\tiny
\setlength{\tabcolsep}{0.1mm}{
\begin{xltabular}{\linewidth}{X}
\# Example 1 \\

\\\texttt{<s> The precise nature of his doubt is not articulated but he aligns himself with the 17th century dissenters who put personal religious conscience before the unifying demands of the Church. This precipitates the move to the North, and determines the whole narrative arc of the novel.
Before we are thrust, along with our central characters, into the smoky North, it is worth lingering a while on this short first stage of the novel. Gaskell does not make a simple thing of the South, as she might have been tempted to do as a clear point of comparison. Instead she offers us two versions of the South, two kinds of imaginings, both of which are then rejected. The drawing-room world of the Shaws, while superficially appealing, is altogether too enervating for the Margaret Hale who is gradually emerging even in these early chapters. Her decided refusal of Lennox is also a refusal of that world. The rural delights of Helstone (in the New Forest) seem initially to offer a simpler, perhaps a truer, version of the South. But that has already been put in doubt by Margaret and Henry Lennox’s rather vexed discussion of it. In London, Henry suggests playfully that it is a ‘village in a tale’, at which Margaret takes umbrage, only to offer instead that it is ‘a village in a poem’ (11). When Henry arrives bang in the middle of that poem, the scene is set for romance: ‘velvety cramoisy roses’ (25), pears plucked from the tree and arranged on a plate of beetroot leaf, and the ‘crimson and amber foliage’ (26) of the deep forest beyond. Yet instead of the completion of the romantic dream, he comes up hard against Margaret’s refusal. Indeed she herself comes up hard against it, and looks back at his proposal somewhat wistfully when she is plunged into her father’s ferment, and briefly longs for the London/Shaw world where nothing ‘called for much decision’.
But if Margaret had accepted Henry – no novel. And besides, he can be kept in mind as a possible future plot line. Gaskell is astute enough to know that challenge makes for a more interesting narrative, and as it turns out, decision is something Margaret is rather good at. In the move to the South, she becomes the adult of the household, her mother declining into frailty, her father exhausted by the consequences of his own conscience. Her growth into her own strength of being is the more convincing because she often quails at what is before her. ‘But the future must be met, however stern and iron it be.’ (55) Thus, after a brief lull and taking of rest at the seaside town of Heston, Margaret and her father make the journey to Milton – the ‘North’ of the novel – where, as she says playfully, ‘“I am overpowered by the discovery of my own genius for management.”’ (57) But the obstacles are real, and the whole family must contend with their much-changed situation:
They were settled in Milton, and must endure smoke and fogs for a season; indeed all other life seemed shut out from them by as thick a fog of circumstance. …
At night when Margaret realised this, she felt inclined to sit down in a stupor of despair. The heavy smoky air hung about her bedroom, which occupied the long narrow projection at the back of the house. The window, placed at the side of the oblong, looked to the blank wall of a similar projection, not above ten feet distant. It loomed through the fog like a great barrier to hope. (62)
We are almost in the world of George Orwell’s The Road to Wigan Pier – a work in fact heavily influenced by nineteenth-century depictions of urban industrialised living conditions. As if to underline their changed life, a letter has come from Edith, full of the delights of married life in Corfu: ‘Edith’s life seemed like the deep vault of blue sky above her, free – utterly free from fleck or cloud’. (62) This leads Margaret to reflect in turn on how, if she had accepted Lennox’s marriage proposal, things might have been different. The omniscient narrative is here able to give much insight into Margaret’s inner thoughts, so that we see her working through these difficult ideas and eventually finding herself clearer and happier: ‘As she realised what might have been, she grew to be thankful for what was.’ (63) If there is a measure of rationalisation in Margaret’s logic here – that too is realistic. }

\\ \hline \\
\# Example 2 \\
\\ \texttt{<s> For example, signals 116 and 118 may be in-phase (I) and quadrature (Q) baseband components of a signal. In the example of FIG. 1B, signals 116 and 118 undergo a zero crossing as they transition from +1 to -1. Signals 116 and 118 are multiplied by signal 120 or signal 120 phase shifted by 90 degrees. Signal 116 is multiplied by a 0 degree shifted version of signal 120. Signal 118 is multiplied by a 90 degree shifted version of signal 120. Resulting signals 122 and 124 represent time-varying complex carrier signals. Note that signals 122 and 124 have envelopes that vary according to the time-varying amplitudes of signals 116 and 118. Further, signals 122 and 124 both undergo phase reversals at the zero crossings of signals 116 and 118. Signals 122 and 124 are summed to result in signal 126. Signal 126 represents a time-varying complex signal. Signal 126 may represent an example input signal into VPA embodiments of the present invention. Additionally, signals 116 and 118 may represent example input signals into VPA embodiments of the present invention.
1.2) Example Generation of Time-Varying Complex Envelope Signals from Constant Envelope Signals
The description in this section generally relates to the operation of step 508 in FIG. 50. FIG. 1C illustrates three examples for the generation of time-varying complex signals from the sum of two or more substantially constant envelope signals. A person skilled in the art will appreciate, however, based on the teachings provided herein that the concepts illustrated in the examples of FIG. 1C can be similarly extended to the case of more than two constant envelope signals.
In example 1 of FIG. 1C, constant envelope signals 132 and 134 are input into phase controller 130. Phase controller 130 manipulates phase components of signals 132 and 134 to generate signals 136 and 138, respectively. Signals 136 and 138 represent substantially constant envelope signals, and are summed to generate signal 140. The phasor representation in FIG. 1C, associated with example 1 illustrates signals 136 and 138 as phasors P136 and P138, respectively. Signal 140 is illustrated as phasor P140. In example 1, P136 and P138 are symmetrically phase shifted by an angle $\phi_1$ relative to a reference signal assumed to be aligned with the real axis of the phasor representation. Correspondingly, time domain signals 136 and 138 are phase shifted in equal amounts but opposite directions relative to the reference signal. Accordingly, P140, which is the sum of P136 and P138, is in-phase with the reference signal.
In example 2 of FIG. 1C, substantially constant envelope signals 132 and 134 are input into phase controller 130. Phase controller 130 manipulates phase components of signals 132 and 134 to generate signals 142 and 144, respectively. Signals 142 and 144 are substantially constant envelope signals, and are summed to generate signal 150. The phasor representation associated with example 2 illustrates signals 142 and 144 as phasors P142 and P144, respectively. Signal 150 is illustrated as phasor P150. In example 2, P142 and P144 are symmetrically phase shifted relative to a reference signal. Accordingly, similar to P140, P150 is also in-phase with the reference signal. P142 and P144, however, are phase shifted by an angle whereby $\phi_2 \neq \phi_1$ relative to the reference signal. P150, as a result, has a different magnitude than P140 of example 1.}\\

\\ \hline \\

\caption{Examples of Randomly Sampled High-Scoring Data Points (Top 10\%).}
\label{tab:vis top10 eg12 ap}
\end{xltabular}}
\end{tcolorbox}

\begin{tcolorbox}[title= Examples of Randomly Sampled High-Scoring Data Points (Top 10\%)]
\tiny
\setlength{\tabcolsep}{0.1mm}{
\begin{xltabular}{\linewidth}{X}
\# Example 3 \\

\\\texttt{<s> Then Eddie owned up and said I took the records and my mummy said what did you do with them Eddie and he said I played cards with them that’s what I done with them where as everybody was playing cards for tuppence and thrupence and he was playing with the records and our Tommy stood and looked at him, I have never forgotten the expression on Tommy’ s face. Eddie was about 15 or 16 then.
When Eddie left school he successfully applied for a job in the Belfast City Council and I remember everybody being very proud because it was difficult to get a job with the Council back then. My mummy came in one day and said I was talking to the foreman about our Eddie and he said he’s a great worker, my mummy was very proud of him. He used to land in for his lunch to my mummy with all the other binmen, she would have to feed them all.
He stayed there until 1968 when he began working at boarding up buildings that had been damaged in the Troubles. As a way to earn an extra bit of money for the family he also worked nights as a barman.
When Eddie got older he was always very particular about his appearance, he always wore a suit, sometimes with shirt or tee shirt, he was always very spick and span. Eddie smoked but he wasn’t a drinker. That’s not to say that he didn’t try it at the beginning but it wasn’t for him, he became a lifelong pioneer and a blood donor. He was also in the Confraternity (which was a sort of prayer group for men) at Clonard Monastery and he loved it; that was his wee place to get away to. He had strong faith.
When Eddie was sixteen he met and fell in love with his future wife Marie. Marie was the love of his life and they courted for six years before getting married in 1962. Five years later their first child was born, quickly followed by three more. Eddie and Marie had three sons – Eamon, Patrick and Ciaran, and one daughter – Brenda.
When they got married they went to live with Marie’s grandmother in Fort Street. But he wanted his own house for himself and Marie and the only way that was going to happen was to get the money together to buy one. My daddy said to him. “look if you’re looking extra money to buy a house go and join the TA its only 2 months a year”. So he went and joined the Territorial Army and the money he was getting he sent it home to Marie. He wasn’t in the TA for very long and had left by the time his first child was born.
When Eddie came home things didn’t work out the way he wanted about the house and things got too much for him, he ended up with a bit of a breakdown. Eventually they got the money for a house in Iveagh Street it was in a bad state of repair but Eddie and Marie fixed it up and made it their home. He suffered with mental health difficulties a couple of times in the early-mid 1960s but that was well behind him by the time of his death.
Eddie just lived for Marie and their kids, he took on a couple of extra jobs, working as a bar man and doing a bit of painting and decorating. He was always ready and willing to drop everything and do something for you. It was just an ordinary family life and he just loved Marie, he idolised her and she could do no wrong in his eyes. All he had time for was work, home and the confraternity. When he did have free time he liked fishing and clay pigeon shooting.
He was content with what he had and he was in his own wee orbit that he owned his house, and provided for his kids he was just happy to be a husband and a daddy. On payday he would give Marie his unopened pay packet, she would then buy him his cigarettes for the week. Not too many men did that in those days.
Eddie was strict in a way too with the kids, I remember Eddie coming to visit me with Ciaran, I had a rocking horse in the living room and Ciaran wanted on it and Eddie said no you’re not going over it doesn’t belong to you, and I looked at Eddie and said let the child go over and get on to the horse I said catch yourself on there is nobody even on it and he went over but he was holding him on it because he maybe would of toppled.}

\\ \hline \\
\# Example 4 \\
\\ 
\texttt{<s> I was interested to see if I would lean closer to earlier poems or later poems since sometimes there can be a significant difference in a poets writing style compared to when they began and ended. Turns ou This.Was.My.Jam Where do I even begin? So the collection is written in reverse chronological- yeah that's right I actually read the introduction to something. I found this particularly interesting because I feel like we often start in the beginning and naturally work our way through their work. I was interested to see if I would lean closer to earlier poems or later poems since sometimes there can be a significant difference in a poets writing style compared to when they began and ended. Turns out I pretty steadily loved it all. I think if I HAD to chose I would lean just slightly closer to the beginning of the collection, but just slightly. That might have a bit of a biases though since Annabelle Leigh is the very first poem we read and it's always been my absolute favorite. Annabelle Leigh aside, I can only imagine what other wonderfully powerful and hauntingly beautiful pieces he could have continued to write had he lived longer. (Internally sobbing) You'll probably notice that there's a lot of reoccurrence with things like the moon, celestial bodies, night, and the evening star- all things I really enjoyed. Also, (and this might quite well be my favorite) Poe has some of the best rhymes. Words that rhymed but weren't your usual rhymes, if you will. For example: departed and brokenhearted, month of June and mystic-moon, dipt in folly and melancholy, Heaven and unforgiven (you gotta twang a little for that one), itself alone and gray tombstone, heart's content and own element. etc. And it doesn't stop there! The the entire language being used is SO GOOD. I'd be reading a poem and then I'd hit a particular line or phrase and just have to a take a moment to say "damn" while the words were absorbed. Some examples of that are "And the silken, sad, uncertain rustling of each purple curtain" (The Raven), I stand amid the roar of a surf-tormented shore (A Dream Within a Dream), With the moon-tints of purple and pearl (Eulalie-A Song), Sound loves to revel in a summer night: Witness the murmur of the gray twilight. (Al Aaraaf Part 2) and "So like you gather in your breath, a portrait taken after death. (Tamberlane) Even the poems that I didn't mark as favorites I still really enjoyed. My least favorite in the collect was Al Aaraaf (both parts), I'm not really sure why I just didn't feel as wow'ed by it. Also, the play that ends the collection I wasn't a huge fan of but I think that just speaks true to the format. Plays are different than poetry. I haven't read any Poe stories for a long time, so I think it would be interesting to see where my enjoyment falls on the prose. But, the poetry is definitely out of the park for me. Something I do find intriguing is that growing up I also thought Poe was just a dark and haunted poet. I think he was in fact haunted, but I don't think (the poetry at least) is as horrific as people usually indicate. In fact, I'm willing to call it beautiful. Beautifully dark, perhaps? Read it, it's perfect.
Mateo – Oct 24, 2020
I did not make this image but this is my review I did not make this image but this is my review
Stephanie Grosse – Sep 23, 2018
This review has been hidden because it contains spoilers. To view it, click here. Simultaneously mysterious and familiar, like the old friend who suddenly astonishes you with his strangeness or the acquaintance whom you are convinced you must have known since childhood. I very much enjoyed the use of onomatopoeia. You will be hypnotised by the sounds (for example "ee", "em" in the summer dream beneath the tamarind tree). Poe has you forever, in "a dream within a dream" Very memorable.A must read for all poetry lovers. Simultaneously mysterious and familiar, like the old friend who suddenly astonishes you with his strangeness or the acquaintance whom you are convinced you must have known since childhood. I very much enjoyed the use of onomatopoeia.}\\

\\ \hline \\

\caption{Examples of Randomly Sampled High-Scoring Data Points (Top 10\%).}
\label{tab:vis top10 eg34 ap}
\end{xltabular}}
\end{tcolorbox}

\begin{tcolorbox}[title= Examples of Randomly Sampled Low-Scoring Data Points (Bottom 10\%)]
\tiny
\setlength{\tabcolsep}{0.1mm}{
\begin{xltabular}{\linewidth}{X}
\# Example 1 \\
\\\texttt{<s>XXXXXXXXXXXXXXXXXXXXXXXXXXXXXXXXXXXXX XXXXXXXXXXXXXXXXXXXXXXXXXXXXXXXXXXXXXXXXXXXXXXXXXXXXXXX  XXXXXXXXXXXXXXXXXXXXXXXXXXXXXXXXXXXXXXXXXXXXXXXXXXXXXXXXXXXXXXXXXXXXXXXXXXXXXXXXXXXXXXXXXXXXXXXXXXXXXXXX XXXXXXXXXXXXXXXXXXXXXXXXXXXXXXXXXXXXXXXXXXXXXXXXXXXXXXXXXXXXXXXXXXXXXXXXXXXXXXXXXXXXXXXXXXXXXXXXXXXXXXXX XXXXXXXXXXXXXXXXXXXXXXXXXXXXXXXXXXXXXXXXXXXXXXXXXXXXXXXXXXXXXXXXXXXXXXXXXXXXXXXXXXXXXXXXXXXXXXXXXXXXXXXX XXXXXXXXXXXXXXXXXXXXXXXXXXXXXXXXXXXXXXXXXXXXXXXXXXXXXXXXXXXXXXXXXXXXXXXXXXXXXXXXXXXXXXXXXXXXXXXXXXXXXXXX XXXXXXXXXXXXXXXXXXXXXXXXXXXXXXXXXXXXXXXXXXXXXXXXXXXXXXXXXXXXXXXXXXXXXXXXXXXXXXXXXXXXXXXXXXXXXXXXXXXXXXXX XXXXXXXXXXXXXXXXXXXXXXXXXXXXXXXXXXXXXXXXXXXXXXXXXXXXXXXXXXXXXXXXXXXXXXXXXXXXXXXXXXXXXXXXXXXXXXXXXXXXXXXX XXXXXXXXXXXXXXXXXXXXXXXXXXXXXXXXXXXXXXXXXXXXXXXXXXXXXXXXXXXXXXXXXXXXXXXXXXXXXXXXXXXXXXXXXXXXXXXXXXXXXXXX XXXXXXXXXXXXXXXXXXXXXXXXXXXXXXXXXXXXXXXXXXXXXXXXXXXXXXXXXXXXXXXXXXXXXXXXXXXXXXXXXXXXXXXXXXXXXXXXXXXXXXXX XXXXXXXXXXXXXXXXXXXXXXXXXXXXXXXXXXXXXXXXXXXXXXXXXXXXXXXXXXXXXXXXXXXXXXXXXXXXXXXXXXXXXXXXXXXXXXXXXXXXXXXX XXXXXXXXXXXXXXXXXXXXXXXXXXXXXXXXXXXXXXXXXXXXXXXXXXXXXXXXXXXXXXXXXXXXXXXXXXXXXXXXXXXXXXXXXXXXXXXXXXXXXXXX
XXXXXXXXXXXXXXXXXXXXXXXXXXXXXXXXXXXXXXXXXXXXXXXXXXXXXXXXXXXXXXXXXXXXXXXXXXXXXXXXXXXXXXXXXXXXXXXXXXXXXXXX XXXXXXXXXXXXXXXXXXXXXXXXXXXXXXXXXXXXXXXXXXXXXXXXXXXXXXXXXXXXXXXXXXXXXXXXXXXXXXXXXXXXXXXXXXXXXXXXXXXXXXXX 
XXXXXXXXXXXXXXXXXXXXXXXXXXXXXXXXXXXXXXXXXXXXXXXXXXXXXXXXXXXXXXXXXXXXXXXXXXXXXXXXXXXXXXXXXXXXXXXXXXXXXXXX
Select a country but NOT a region
All you need to know about the Indian Defence Forces!}

\\ \hline \\
\# Example 2 \\ 
\\\texttt{<s>XXXXXXXXXXXXXXXXXXXXXXXXXXXXXXXXXXXXX XXXXXXXXXXXXXXXXXXXXXXXXXXXXXXXXXXXXXXXXXXXXXXXXXXXXXXX  XXXXXXXXXXXXXXXXXXXXXXXXXXXXXXXXXXXXXXXXXXXXXXXXXXXXXXXXXXXXXXXXXXXXXXXXXXXXXXXXXXXXXXXXXXXXXXXXXXXXXXXX XXXXXXXXXXXXXXXXXXXXXXXXXXXXXXXXXXXXXXXXXXXXXXXXXXXXXXXXXXXXXXXXXXXXXXXXXXXXXXXXXXXXXXXXXXXXXXXXXXXXXXXX XXXXXXXXXXXXXXXXXXXXXXXXXXXXXXXXXXXXXXXXXXXXXXXXXXXXXXXXXXXXXXXXXXXXXXXXXXXXXXXXXXXXXXXXXXXXXXXXXXXXXXXX XXXXXXXXXXXXXXXXXXXXXXXXXXXXXXXXXXXXXXXXXXXXXXXXXXXXXXXXXXXXXXXXXXXXXXXXXXXXXXXXXXXXXXXXXXXXXXXXXXXXXXXX XXXXXXXXXXXXXXXXXXXXXXXXXXXXXXXXXXXXXXXXXXXXXXXXXXXXXXXXXXXXXXXXXXXXXXXXXXXXXXXXXXXXXXXXXXXXXXXXXXXXXXXX XXXXXXXXXXXXXXXXXXXXXXXXXXXXXXXXXXXXXXXXXXXXXXXXXXXXXXXXXXXXXXXXXXXXXXXXXXXXXXXXXXXXXXXXXXXXXXXXXXXXXXXX XXXXXXXXXXXXXXXXXXXXXXXXXXXXXXXXXXXXXXXXXXXXXXXXXXXXXXXXXXXXXXXXXXXXXXXXXXXXXXXXXXXXXXXXXXXXXXXXXXXXXXXX XXXXXXXXXXXXXXXXXXXXXXXXXXXXXXXXXXXXXXXXXXXXXXXXXXXXXXXXXXXXXXXXXXXXXXXXXXXXXXXXXXXXXXXXXXXXXXXXXXXXXXXX XXXXXXXXXXXXXXXXXXXXXXXXXXXXXXXXXXXXXXXXXXXXXXXXXXXXXXXXXXXXXXXXXXXXXXXXXXXXXXXXXXXXXXXXXXXXXXXXXXXXXXXX XXXXXXXXXXXXXXXXXXXXXXXXXXXXXXXXXXXXXXXXXXXXXXXXXXXXXXXXXXXXXXXXXXXXXXXXXXXXXXXXXXXXXXXXXXXXXXXXXXXXXXXX
XXXXXXXXXXXXXXXXXXXXXXXXXXXXXXXXXXXXXXXXXXXXXXXXXXXXXXXXXXXXXXXXXXXXXXXXXXXXXXXXXXXXXXXXXXXXXXXXXXXXXXXX XXXXXXXXXXXXXXXXXXXXXXXXXXXXXXXXXXXXXXXXXXXXXXXXXXXXXXXXXXXXXXXXXXXXXXXXXXXXXXXXXXXXXXXXXXXXXXXXXXXXXXXX 
XXXXXXXXXXXXXXXXXXXXXXXXXXXXXXXXXXXXXXXXXXXXXXXXXXXXXXXXXXXXXXXXXXXXXXXXXXXXXXXXXXXXXXXXXXXXXXXXXXXXXXXX
Select a country but NOT a region
A site focusing on Australian Modelling with galleries, articles and discussion forums.}

\\ \hline \\
\# Example 3 \\ 
\\\texttt{
<s>USA, Liberia
USA, Lithuania, Italy
USA, Luxembourg
USA, Luxembourg, UK
USA, Malaysia
USA, Malta, France, UK
USA, Malta, UK
USA, Mexico
USA, Mexico, Australia
USA, Mexico, Australia, Canada
USA, Mexico, Canada
USA, Mexico, Canada, Germany
USA, Mexico, Germany
USA, Mexico, Hong Kong
USA, Mexico, Japan
USA, Mexico, Spain
USA, Mexico, UK
USA, Mexico, United Arab Emirates
USA, Monaco
USA, Monaco, Morocco
USA, Morocco
USA, Morocco, Spain, UK
USA, Morocco, Switzerland
USA, Myanmar
USA, Netherlands
USA, Netherlands, France
USA, Netherlands, Germany, France, Austria
USA, Netherlands, South Africa
USA, Netherlands, UK
USA, Netherlands, UK, Denmark
USA, New Zealand
USA, New Zealand, Canada, Israel, Japan, Nigeria
USA, New Zealand, Germany
USA, New Zealand, Japan
USA, New Zealand, South Africa, UK, Lithuania
USA, New Zealand, UK
USA, Nicaragua
USA, Nigeria
USA, Norway
USA, Pakistan
USA, Panama, Argentina
USA, Panama, Japan, Canada
USA, Panama, Mexico
USA, Peru
USA, Philippines
USA, Philippines, Puerto Rico
USA, Philippines, Taiwan, South Korea, China, Canada
USA, Poland
USA, Poland, Slovenia, Czech Republic, UK
USA, Portugal
USA, Portugal, France
USA, Puerto Rico
USA, Qatar
USA, Romania
USA, Romania, Canada
USA, Romania, France, Italy, Germany
USA, Romania, Germany
USA, Romania, Iceland
USA, Romania, UK
USA, Russia
USA, Russia, Hungary
USA, Russia, UK
USA, Saudi Arabia
USA, Senegal
USA, Serbia
USA, Serbia, Canada
USA, Singapore
USA, Singapore, Taiwan
USA, Slovakia
USA, Slovakia, China
USA, South Africa
USA, South Africa, Germany
USA, South Africa, India
USA, South Africa, Italy
USA, South Africa, Zambia, Germany
USA, South Korea
USA, South Korea, Australia
USA, South Korea, India
USA, South Korea, Japan
USA, South Korea, Singapore
USA, South Korea, Singapore, Russia, Malaysia, Kazakhstan, Taiwan, Hong Kong, Japan, China, India, Syria, Iran, Egypt, Pakistan
USA, South Korea, Spain
}

\\ \hline \\
\# Example 4 \\ 
\\\texttt{
Phillip L. Horrell v. David Gomez, Warden, No. 20-5306
Ganaa Otgoo v. Illinois, No. 20-5109
Phillip Hartsfield v. Stepanie Dorethy, Warden, No. 19-1473
Anthony Jackson v. Supreme Court of Illinois, No. 19-8665
David Beverly v. Illinois, No. 19-8502
Lamont Dantzler v. Illinois, No. 19-8448
Joh-ner Taylor Wilson v. Illinois, No. 19-8437
Herbert Burgess v. Illinois, No. 19-8379
Joseph M. Coffman v. Illinois, No. 19-8391
Timothy J. McVay v. Illinois, No. 19-8304
Kenneth Durant v. Frank Lawrence, Warden, No. 19-7967
Seth A. Weaver v. Illinois, No. 19-7823
Lyarron T. Emers v. Illinois, No. 19-7759
Bethany Austin v. Illinois, No. 19-1029
Anthony Allen v. Illinois, No. 19-7633
Kenin L. Edwards v. Michael L. Atterberry, et al., No. 19-965
Pablo Rodriguez-Palomino v. Illinois, No. 19-7273
Christopher L. Croom v. Illinois, No. 19-7237
Tony Robinson v. Illinois, No. 19-7226
Lazaro Zapata v. Illinois, No. 19-7264
Fernando Oliveros v. Illinois, No. 19-7141
Kevin Dameron v. Illinois, No. 19-6945
Peter Gakuba v. Michelle Neese, No. 19-6543
Richard Kalinowski v. Illinois, No. 19-6368
Rafael Alvarado v. Frank Lawrence, Warden, No. 19-6347
Chad M. Cutler v. Illinois, No. 19-6150
Hezekiah Whitfield v. Deanna Brookhart, Warden, No. 19-6051
Lorenzo Davis, Jr. v. Illinois, No. 19-5831
Chadwick N. Barner v. Illinois, No. 19-5655
Charles Donelson v. Q. Tanner, et al., No. 19-5397
Robert Curry v. Illinois, No. 19-5366
Andrew Condon v. Illinois, No. 19-5349
Keith Talbert v. Illinois, No. 18-9768
Juan Rodriguez v. Illinois, No. 18-9759
Miguel Alcantar v. Illinois, No. 18-1548
Gregory Rayford v. Illinois, No. 18-9612
Irving Madden v. Michael Melvin, Warden, No. 18-9474
Denzel Pittman v. Illinois, No. 18-9451
Jesus Cotto v. Jacqueline Lashbrook, Warden, No. 18-9116
Pierre Montanez v. Ursula Walowski, No. 18-9101
Russell Frey v. Illinois, No. 18-9120
Peter Gakuba v. Illinois, No. 18-9041
Jose Cobian v. Illinois, No. 18-8963
Derrick Redmond v. Illinois, No. 18-8808
Jennifer N. Nere v. Illinois, No. 18-8625
Gerald W. Long v. Illinois, No. 18-8577
Willie White v.
}
\\

\\ \hline \\
\caption{Examples of Randomly Sampled Low-Scoring Data Points (Bottom 10\%).}
\end{xltabular}}
\label{tab:vis bottom10 ap}
\end{tcolorbox}

%% file: Tabs/exp_tab1_ap.tex
\begin{table}[htbp!]
\vspace{-0.3cm}
\centering
\scriptsize

\caption{
Efficacy results on different downstream benchmarks.
Ours means applying LQS for data scoring and Folding for data ordering within the DELT paradigm.
}
\vspace{-0.5cm}
\begin{subtable}
\centering
\caption*{(a) Results (\%) for 1B-token data across model sizes (160M, 470M, 1B).}
\label{tab: model size}
\begin{tabular}{c|cccccccc|c}
\toprule
\multicolumn{1}{l}{} & ARC-c            & ARC-e            & HS               & LAMB             & OBQA             & PIQA             & SciQ             & Wino             & Avg.             \\
\midrule
\multicolumn{1}{l}{} & \multicolumn{9}{c}{Model size = 160M}                            \\ \midrule
Conventional         & 21.27          & 34.32          & 27.85          & 20.25          & 24.40          & 55.19 & 56.93          & 50.72          & 36.37          \\
KenLM                & \textbf{21.93} & 33.96 & 28.09 & 20.69 & 25.20 & 54.79          & 56.20 & 50.59 & 36.43 \\
PDS                  & 21.84          & 35.02          & 27.61          & 19.93          & 24.80          & \textbf{56.23} & 59.00          & 51.38          & 37.01          \\
Ours                 & 21.59          & \textbf{36.07} & \textbf{28.41} & \textbf{23.79} & \textbf{25.60} & 56.37          & \textbf{59.80} & \textbf{53.04} & \textbf{38.08} \\
\midrule
\multicolumn{1}{l}{} & \multicolumn{9}{c}{Model size = 470M}                            \\ \midrule
Conventional         & 21.16          & 34.91          & 28.11          & 21.88          & 23.90          & 56.07          & 58.75          & 50.04          & 36.85          \\
KenLM                & \textbf{22.35}          & 34.85          & 28.05          & 20.51          & 25.00          & 55.17          & 56.60           & 50.04          & 36.57                \\
PDS                  & 22.10          & 33.04          & 27.84          & 21.25          & 24.80          & 56.96          & 59.80           & 51.85          & 37.23                \\
Ours                 & 22.33 & \textbf{35.88} & \textbf{28.45} & \textbf{23.26} & \textbf{26.60} & \textbf{57.20} & \textbf{60.10} & \textbf{52.81} & \textbf{38.33}       \\
\midrule
\multicolumn{1}{l}{} & \multicolumn{9}{c}{Model size = 1B}         
\\ \midrule
Conventional    &20.58   & 36.12   & 28.32      & 23.56     & 25.00     & 56.49     & 60.05     & 52.07     & 37.77 \\
KenLM           &21.67   & 35.86   & 28.76      & 23.46     & \textbf{26.80}      & 56.58     & 59.00     & 49.88     & 37.75   \\
PDS             &22.10    & 35.56   & 28.20       & 23.56     & 26.40      & 56.37     & 60.50      & 50.67     & 37.92   \\
Ours                 & \textbf{22.76} & \textbf{37.95} & \textbf{29.95} & \textbf{26.38} & 26.00 & \textbf{58.07} & \textbf{60.90} & \textbf{51.28} & \textbf{39.17} \\
\bottomrule
\end{tabular}
\end{subtable}
\begin{subtable}
\centering
\caption*{(b) Results (\%) for 160M model across data sizes (10B, 50B).}
\label{tab: data size}
\begin{tabular}{c|cccccccc|c}
\toprule
\multicolumn{1}{l}{} & ARC-c            & ARC-e            & HS               & LAMB             & OBQA             & PIQA             & SciQ             & Wino             & Avg.             \\
\midrule
\multicolumn{1}{l}{} & \multicolumn{9}{c}{Data size = 10B tokens}                                                                                                                               \\
\midrule
Conventional         & 22.82          & 38.51          & 30.72          & 30.40          & 25.70          & 57.32          & 64.90          & 51.54          & 40.24          \\
KenLM                & 22.78  & 37.92  & 30.54  & 29.98  & 25.60  & 57.29  & 66.00   & 52.80  & 40.36 \\
PDS                  & 22.70  & 39.35 & 30.73 & 31.85 & 27.20 & 56.04 & 64.90   & \textbf{52.88}   & 40.71     \\
Ours                 & \textbf{24.38} & \textbf{39.80} & \textbf{31.64} & \textbf{32.98} & \textbf{27.21} & \textbf{58.56} & \textbf{66.70} & 51.67 & \textbf{41.62} \\
\midrule
\multicolumn{1}{l}{} & \multicolumn{9}{c}{Data size = 50B tokens}                                                                                                                               \\
\midrule
Conventional         & 24.06 & \textbf{41.88} & 32.05 & 33.79 & 26.80 & 58.11 & \textbf{69.00} & 51.93 & 42.20          \\
KenLM                & 23.74 & 40.14 & 32.10 & 35.13 & 28.41 & 58.15 & 67.52 & 51.71 & 42.11          \\
PDS                  & 24.57 & 41.37 & 32.44 & 35.36 & 29.20 & \textbf{59.25} & 68.10 & 50.83 & 42.64 \\
Ours                 & \textbf{24.65} & 41.07 & \textbf{33.00}    & \textbf{36.07} & \textbf{29.30} & 59.10  & 68.40 & \textbf{52.67} & \textbf{43.03}         \\
\bottomrule
\end{tabular}
\end{subtable}
\label{tab: appendix model data size}
\end{table}

%% file: Tabs/exp_tab2_ap.tex

\begin{table}[hbtp!]
\vspace{-0.1cm}
\centering
\scriptsize
\setlength{\tabcolsep}{5pt}
\caption{
Efficiency results of different methods.
The selection methods report the highest scores across all selection ratios.
The best scores for each model size are highlighted in \textbf{bold}, while the second-best scores are shown in \textbf{\textit{italic bold}}.
}
\begin{tabular}{p{1.2cm}|p{1.2cm}p{0.6cm}p{0.7cm}|cccccccc|c}
\toprule
Pipeline                         & Scoring & Selection & Ordering & ARC-c            & ARC-e            & HS               & LAMB             & OBQA             & PIQA             & SciQ             & Wino             & Avg.             \\
\midrule
Conventional                     & -       & -         & -        & 21.27          & 34.32          & 27.85          & 20.25          & 24.40          & 55.19          & 56.93          & 50.72          & 36.37          \\
\midrule
\multirow{2}{*}{Efficiency} & KenLM\citep{heafield-2011-kenlm} & \checkmark       & -        & 21.42 & 34.34 & 27.76 & 20.84 & 25.00 & \textit{\textbf{56.31}} & 54.30 & 51.07 & 36.38          \\
                                 & PDS\citep{gu2024data}     & \checkmark       & -        & 21.84          & \textit{\textbf{35.02}}          & 27.61          & 19.93          & 24.80          & 56.23          & \textbf{\textit{59.00}}          & 51.38          & 37.01          \\
                            & LQS (Ours) & \checkmark & - & \textbf{22.18} & 34.09 & \textit{\textbf{27.80}} & \textit{\textbf{21.02}} & \textit{\textbf{25.20}} & 55.98 & \textit{\textbf{59.00}} & \textit{\textbf{51.85}} & \textit{\textbf{37.14}} \\
\midrule
DELT       & LQS (Ours) & \checkmark       & Folding  & 21.59          & \textbf{36.07} & \textbf{28.41} & \textbf{23.79} & \textbf{25.60}          & \textbf{56.37}          & \textbf{59.80} & \textbf{53.04} & \textbf{38.08}           \\ 
\bottomrule
\end{tabular}
\label{tab:efficiencyvsefficacy_ap}
\end{table}

%% file: Tabs/exp_epochs_ap.tex
\begin{table}[htbp!]
\centering
\scriptsize
\caption{Results on OLMo for the different epochs.}
\label{tab:ep_ap}
\begin{tabular}{c|c|cccccccc|c}
\toprule

Epoch              & \multicolumn{1}{l}{} & ARC-c          & ARC-e          & HS             & LAMB           & OBQA          & PIQA           & SciQ          & Wino           & Avg.           \\
\midrule
\multirow{2}{*}{1} & Conventional         & 21.27          & 34.32          & 27.85          & 20.25          & 24.40          & 55.19          & 56.93         & 50.72          & 36.37          \\
                   & DELT (Ours)          & \textbf{21.59} & \textbf{36.07} & \textbf{28.41} & \textbf{23.79} & \textbf{25.60} & \textbf{56.37} & \textbf{59.80} & \textbf{53.04} & \textbf{38.08} \\ 
\midrule
\multirow{2}{*}{2} & Conventional         & 21.93          & 36.20           & \textbf{29.18} & 25.93          & 23.00            & \textbf{56.86} & \textbf{61.20} & 50.99          & 38.16          \\ 
                   & DELT (Ours)          & \textbf{22.35} & \textbf{36.41} & 28.28          & \textbf{27.63} & \textbf{26.80} & 56.47          & 61.00            & \textbf{51.22} & \textbf{38.77} \\
\midrule
\multirow{2}{*}{3} & Conventional         & 21.35          & 35.78          & 28.76          & 27.14          & \textbf{26.20} & \textbf{56.51} & \textbf{62.80} & 49.51          & 38.51          \\
                   & DELT (Ours)          & \textbf{22.44} & \textbf{36.95} & \textbf{29.41} & \textbf{29.09} & 24.80          & 56.20           & 62.30          & \textbf{51.62} & \textbf{39.10}  \\
\midrule
\multirow{2}{*}{4} & Conventional         & 21.10           & 35.99          & 28.97          & 27.51          & 27.20          & 55.69          & 61.80          & 49.28          & 38.44          \\
                   & DELT (Ours)          & \textbf{22.53} & \textbf{38.05} & \textbf{29.78} & \textbf{29.58} & \textbf{26.40} & \textbf{57.34} & \textbf{63.90} & \textbf{51.85} & \textbf{39.93} \\
\midrule
\multirow{2}{*}{5} & Conventional         & 20.59          & 37.55          & 29.31          & 28.05          & \textbf{27.00}   & 57.11          & 61.20          & \textbf{50.54} & 38.92          \\
                   & DELT (Ours)          & \textbf{22.87} & \textbf{38.05} & \textbf{30.01} & \textbf{30.08} & 26.80          & \textbf{58.16} & \textbf{64.10} & 49.80           & \textbf{39.98} \\

\bottomrule
\end{tabular}
\end{table}

%% file: Tabs/exp_ablation_l.tex
\begin{table}[htbp!]
\centering
\scriptsize
\caption{Effect of the fold layer $L$.
$L = -$ represents the conventional method, which is three times the random average results.
When $L=1$, the ordering method reduces to curriculum learning.
}
\label{tab:ablation l ap}
\begin{tabular}{c|cccccccc|c}
\toprule
$L$ & ARC-c          & ARC-e          & HS             & LAMB           & OBQA          & PIQA           & SciQ          & Wino           & Avg.           \\
\midrule
- & 21.27	&34.32	&27.85	&20.25	&24.40	&55.19	&56.93	&50.72	&36.37           \\
\midrule
1 & 22.18          & 35.40           & 28.01          & 23.48          & 23.80          & 55.60           & 56.80          & 51.07          & 37.04          \\
2 & 21.57          & 34.26          & 28.34          & 23.29          & 25.80          & 55.88          & 58.70          & 49.80           & 37.21          \\
3 & 21.59          & \textbf{36.07} & \textbf{28.41} & \textbf{23.79} & 25.60          & 56.37          & \textbf{59.80} & \textbf{53.04} & \textbf{38.08} \\
4 & 22.83          & 34.98          & 28.50           & 22.35          & 24.90          & \textbf{56.67} & 59.80          & 50.10           & 37.52          \\
5 & \textbf{22.91} & 35.57          & 28.16          & 22.85          & \textbf{26.70} & 55.41          & 57.30          & 52.08          & 37.62    \\
\bottomrule
\end{tabular}
\end{table}

%% file: neurips_2025.bbl
\begin{thebibliography}{10}

\bibitem{ouyang2022training}
Long Ouyang, Jeffrey Wu, Xu~Jiang, Diogo Almeida, Carroll Wainwright, Pamela Mishkin, Chong Zhang, Sandhini Agarwal, Katarina Slama, Alex Ray, et~al.
\newblock Training language models to follow instructions with human feedback.
\newblock {\em Advances in neural information processing systems}, 35:27730--27744, 2022.

\bibitem{achiam2023gpt}
Josh Achiam, Steven Adler, Sandhini Agarwal, Lama Ahmad, Ilge Akkaya, Florencia~Leoni Aleman, Diogo Almeida, Janko Altenschmidt, Sam Altman, Shyamal Anadkat, et~al.
\newblock Gpt-4 technical report.
\newblock {\em arXiv preprint arXiv:2303.08774}, 2023.

\bibitem{dubey2024llama}
Abhimanyu Dubey, Abhinav Jauhri, Abhinav Pandey, Abhishek Kadian, Ahmad Al-Dahle, Aiesha Letman, Akhil Mathur, Alan Schelten, Amy Yang, Angela Fan, et~al.
\newblock The llama 3 herd of models.
\newblock {\em arXiv preprint arXiv:2407.21783}, 2024.

\bibitem{hirschberg2015advances}
Julia Hirschberg and Christopher~D Manning.
\newblock Advances in natural language processing.
\newblock {\em Science}, 349(6245):261--266, 2015.

\bibitem{gunasekaran2023exploring}
Karthick~Prasad Gunasekaran.
\newblock Exploring sentiment analysis techniques in natural language processing: A comprehensive review.
\newblock {\em arXiv preprint arXiv:2305.14842}, 2023.

\bibitem{yu2024natural}
Fei Yu, Hongbo Zhang, Prayag Tiwari, and Benyou Wang.
\newblock Natural language reasoning, a survey.
\newblock {\em ACM Computing Surveys}, 56(12):1--39, 2024.

\bibitem{kusal2022ai}
Sheetal Kusal, Shruti Patil, Jyoti Choudrie, Ketan Kotecha, Sashikala Mishra, and Ajith Abraham.
\newblock Ai-based conversational agents: a scoping review from technologies to future directions.
\newblock {\em IEEE Access}, 10:92337--92356, 2022.

\bibitem{albalak2024dssurvey}
Alon Albalak, Yanai Elazar, Sang~Michael Xie, Shayne Longpre, Nathan Lambert, Xinyi Wang, Niklas Muennighoff, Bairu Hou, Liangming Pan, Haewon Jeong, et~al.
\newblock A survey on data selection for language models.
\newblock {\em arXiv preprint arXiv:2402.16827}, 2024.

\bibitem{xie2023data}
Sang~Michael Xie, Shibani Santurkar, Tengyu Ma, and Percy~S Liang.
\newblock Data selection for language models via importance resampling.
\newblock {\em Advances in Neural Information Processing Systems}, 36:34201--34227, 2023.

\bibitem{gu2024data}
Yuxian Gu, Li~Dong, Hongning Wang, Yaru Hao, Qingxiu Dong, Furu Wei, and Minlie Huang.
\newblock Data selection via optimal control for language models.
\newblock {\em arXiv preprint arXiv:2410.07064}, 2024.

\bibitem{campos2021curriculum}
Daniel Campos.
\newblock Curriculum learning for language modeling.
\newblock {\em arXiv preprint arXiv:2108.02170}, 2021.

\bibitem{wang2021survey}
Xin Wang, Yudong Chen, and Wenwu Zhu.
\newblock A survey on curriculum learning.
\newblock {\em IEEE transactions on pattern analysis and machine intelligence}, 44(9):4555--4576, 2021.

\bibitem{openAI4o2024}
OpenAI.
\newblock {hello-gpt-4o}.
\newblock (2024).

\bibitem{team2023gemini}
Gemini Team, Rohan Anil, Sebastian Borgeaud, Jean-Baptiste Alayrac, Jiahui Yu, Radu Soricut, Johan Schalkwyk, Andrew~M Dai, Anja Hauth, Katie Millican, et~al.
\newblock Gemini: a family of highly capable multimodal models.
\newblock {\em arXiv preprint arXiv:2312.11805}, 2023.

\bibitem{cho2014learning}
Kyunghyun Cho, Bart Van~Merri{\"e}nboer, Caglar Gulcehre, Dzmitry Bahdanau, Fethi Bougares, Holger Schwenk, and Yoshua Bengio.
\newblock Learning phrase representations using rnn encoder-decoder for statistical machine translation.
\newblock {\em arXiv preprint arXiv:1406.1078}, 2014.

\bibitem{hochreiter1997long}
Sepp Hochreiter and J{\"u}rgen Schmidhuber.
\newblock Long short-term memory.
\newblock {\em Neural computation}, 9(8):1735--1780, 1997.

\bibitem{kaplan2020scaling}
Jared Kaplan, Sam McCandlish, Tom Henighan, Tom~B Brown, Benjamin Chess, Rewon Child, Scott Gray, Alec Radford, Jeffrey Wu, and Dario Amodei.
\newblock Scaling laws for neural language models.
\newblock {\em arXiv preprint arXiv:2001.08361}, 2020.

\bibitem{goyal2024scaling}
Sachin Goyal, Pratyush Maini, Zachary~C Lipton, Aditi Raghunathan, and J~Zico Kolter.
\newblock Scaling laws for data filtering--data curation cannot be compute agnostic.
\newblock In {\em Proceedings of the IEEE/CVF Conference on Computer Vision and Pattern Recognition}, pages 22702--22711, 2024.

\bibitem{heafield-2011-kenlm}
Kenneth Heafield.
\newblock {K}en{LM}: Faster and smaller language model queries.
\newblock In Chris Callison-Burch, Philipp Koehn, Christof Monz, and Omar~F. Zaidan, editors, {\em Proceedings of the Sixth Workshop on Statistical Machine Translation}, pages 187--197, Edinburgh, Scotland, July 2011. Association for Computational Linguistics.

\bibitem{anthropic2024}
Anthropic.
\newblock { Claude 3 haiku: our fastest model yet.}
\newblock (2024).

\bibitem{yang2007paml}
Ziheng Yang.
\newblock Paml 4: phylogenetic analysis by maximum likelihood.
\newblock {\em Molecular biology and evolution}, 24(8):1586--1591, 2007.

\bibitem{cc:Rana:2010:Common-Crawl-open-web-scale-crawl}
Ahad Rana.
\newblock Common crawl – building an open web-scale crawl using hadoop, 2010.

\bibitem{Gutenberg2004}
Michael Hart.
\newblock Project gutenberg, 2004.

\bibitem{nikolenko2021synthetic}
Sergey~I Nikolenko et~al.
\newblock {\em Synthetic data for deep learning}, volume 174.
\newblock Springer, 2021.

\bibitem{kabadayi2006virtual}
Sanem Kabadayi, Adam Pridgen, and Christine Julien.
\newblock Virtual sensors: Abstracting data from physical sensors.
\newblock In {\em 2006 International Symposium on a World of Wireless, Mobile and Multimedia Networks (WoWMoM'06)}, pages 6--pp. IEEE, 2006.

\bibitem{raffel2020exploring}
Colin Raffel, Noam Shazeer, Adam Roberts, Katherine Lee, Sharan Narang, Michael Matena, Yanqi Zhou, Wei Li, and Peter~J Liu.
\newblock Exploring the limits of transfer learning with a unified text-to-text transformer.
\newblock {\em Journal of machine learning research}, 21(140):1--67, 2020.

\bibitem{penedo2023refinedweb}
Guilherme Penedo, Quentin Malartic, Daniel Hesslow, Ruxandra Cojocaru, Hamza Alobeidli, Alessandro Cappelli, Baptiste Pannier, Ebtesam Almazrouei, and Julien Launay.
\newblock The refinedweb dataset for falcon llm: Outperforming curated corpora with web data only.
\newblock {\em Advances in Neural Information Processing Systems}, 36:79155--79172, 2023.

\bibitem{weber2024redpajama}
Maurice Weber, Dan Fu, Quentin Anthony, Yonatan Oren, Shane Adams, Anton Alexandrov, Xiaozhong Lyu, Huu Nguyen, Xiaozhe Yao, Virginia Adams, et~al.
\newblock Redpajama: an open dataset for training large language models.
\newblock {\em Advances in neural information processing systems}, 37:116462--116492, 2024.

\bibitem{chang2024redstone}
Yaoyao Chang, Lei Cui, Li~Dong, Shaohan Huang, Yangyu Huang, Yupan Huang, Scarlett Li, Tengchao Lv, Shuming Ma, Qinzheng Sun, et~al.
\newblock Redstone: Curating general, code, math, and qa data for large language models.
\newblock {\em arXiv preprint arXiv:2412.03398}, 2024.

\bibitem{yu2024mates}
Zichun Yu, Spandan Das, and Chenyan Xiong.
\newblock Mates: Model-aware data selection for efficient pretraining with data influence models.
\newblock {\em Advances in Neural Information Processing Systems}, 2024.

\bibitem{daitfdp}
Yalun Dai, Lingao Xiao, Ivor Tsang, and Yang He.
\newblock Training-free dataset pruning for instance segmentation.
\newblock In {\em The Thirteenth International Conference on Learning Representations}.

\bibitem{zhao2021p}
QiHao Zhao, Wei Hu, Yangyu Huang, and Fan Zhang.
\newblock P-diff+: Improving learning classifier with noisy labels by noisy negative learning loss.
\newblock {\em Neural Networks}, 144:1--10, 2021.

\bibitem{hu2021p}
Wei Hu, QiHao Zhao, Yangyu Huang, and Fan Zhang.
\newblock P-diff: Learning classifier with noisy labels based on probability difference distributions.
\newblock In {\em 2020 25th International Conference on Pattern Recognition (ICPR)}, pages 1882--1889. IEEE, 2021.

\bibitem{abbas2023semdedup}
Amro Abbas, Kushal Tirumala, D{\'a}niel Simig, Surya Ganguli, and Ari~S Morcos.
\newblock Semdedup: Data-efficient learning at web-scale through semantic deduplication.
\newblock {\em arXiv preprint arXiv:2303.09540}, 2023.

\bibitem{tirumala2023d4}
Kushal Tirumala, Daniel Simig, Armen Aghajanyan, and Ari Morcos.
\newblock D4: Improving llm pretraining via document de-duplication and diversification.
\newblock {\em Advances in Neural Information Processing Systems}, 36:53983--53995, 2023.

\bibitem{kim2024strategic}
Jisu Kim and Juhwan Lee.
\newblock Strategic data ordering: Enhancing large language model performance through curriculum learning.
\newblock {\em arXiv preprint arXiv:2405.07490}, 2024.

\bibitem{chang2021does}
Ernie Chang, Hui-Syuan Yeh, and Vera Demberg.
\newblock Does the order of training samples matter? improving neural data-to-text generation with curriculum learning.
\newblock {\em arXiv preprint arXiv:2102.03554}, 2021.

\bibitem{doremi}
Sang~Michael Xie, Hieu Pham, Xuanyi Dong, Nan Du, Hanxiao Liu, Yifeng Lu, Percy~S Liang, Quoc~V Le, Tengyu Ma, and Adams~Wei Yu.
\newblock {DoReMi}: Optimizing data mixtures speeds up language model pretraining.
\newblock In {\em Proceedings of NeurIPS}, 2024.

\bibitem{lima}
Chunting Zhou, Pengfei Liu, Puxin Xu, Srinivasan Iyer, Jiao Sun, Yuning Mao, Xuezhe Ma, Avia Efrat, Ping Yu, Lili Yu, et~al.
\newblock Lima: Less is more for alignment.
\newblock In {\em Proceedings of NeurIPS}, 2024.

\bibitem{paster2023openwebmath}
Keiran Paster, Marco~Dos Santos, Zhangir Azerbayev, and Jimmy Ba.
\newblock Openwebmath: An open dataset of high-quality mathematical web text, 2023.

\bibitem{zheng2021minif2f}
Kunhao Zheng, Jesse~Michael Han, and Stanislas Polu.
\newblock Minif2f: a cross-system benchmark for formal olympiad-level mathematics.
\newblock {\em arXiv preprint arXiv:2109.00110}, 2021.

\bibitem{lozhkov2024starcoder}
Anton Lozhkov, Raymond Li, Loubna~Ben Allal, Federico Cassano, Joel Lamy-Poirier, Nouamane Tazi, Ao~Tang, Dmytro Pykhtar, Jiawei Liu, Yuxiang Wei, et~al.
\newblock Starcoder 2 and the stack v2: The next generation.
\newblock {\em arXiv preprint arXiv:2402.19173}, 2024.

\bibitem{wang2025epicoder}
Yaoxiang Wang, Haoling Li, Xin Zhang, Jie Wu, Xiao Liu, Wenxiang Hu, Zhongxin Guo, Yangyu Huang, Ying Xin, Yujiu Yang, et~al.
\newblock Epicoder: Encompassing diversity and complexity in code generation.
\newblock {\em arXiv preprint arXiv:2501.04694}, 2025.

\bibitem{mistral}
Albert~Q Jiang, Alexandre Sablayrolles, Arthur Mensch, Chris Bamford, Devendra~Singh Chaplot, Diego de~las Casas, Florian Bressand, Gianna Lengyel, Guillaume Lample, Lucile Saulnier, et~al.
\newblock Mistral 7b.
\newblock {\em arXiv preprint arXiv:2310.06825}, 2023.

\bibitem{bai2023qwen}
Jinze Bai, Shuai Bai, Yunfei Chu, Zeyu Cui, Kai Dang, Xiaodong Deng, Yang Fan, Wenbin Ge, Yu~Han, Fei Huang, et~al.
\newblock Qwen technical report.
\newblock {\em arXiv preprint arXiv:2309.16609}, 2023.

\bibitem{olmo}
Dirk Groeneveld, Iz~Beltagy, Pete Walsh, Akshita Bhagia, Rodney Kinney, Oyvind Tafjord, Ananya~Harsh Jha, Hamish Ivison, Ian Magnusson, Yizhong Wang, et~al.
\newblock {OLMo}: Accelerating the science of language models.
\newblock {\em arXiv preprint arXiv:2402.00838}, 2024.

\bibitem{hella_swag}
Rowan Zellers, Ari Holtzman, Yonatan Bisk, Ali Farhadi, and Yejin Choi.
\newblock Hellaswag: Can a machine really finish your sentence?
\newblock In {\em Proceedings of ACL}, 2019.

\bibitem{winogrande}
Hector Levesque, Ernest Davis, and Leora Morgenstern.
\newblock The winograd schema challenge.
\newblock In {\em Proceedings of KR}, 2012.

\bibitem{lambada}
Denis Paperno, Germ{\'a}n Kruszewski, Angeliki Lazaridou, Ngoc-Quan Pham, Raffaella Bernardi, Sandro Pezzelle, Marco Baroni, Gemma Boleda, and Raquel Fern{\'a}ndez.
\newblock The lambada dataset: Word prediction requiring a broad discourse context.
\newblock In {\em Proceedings of ACL}, 2016.

\bibitem{openbookqa}
Todor Mihaylov, Peter Clark, Tushar Khot, and Ashish Sabharwal.
\newblock Can a suit of armor conduct electricity? a new dataset for open book question answering.
\newblock In {\em Proceedings of EMNLP}, 2018.

\bibitem{arc}
Peter Clark, Isaac Cowhey, Oren Etzioni, Tushar Khot, Ashish Sabharwal, Carissa Schoenick, and Oyvind Tafjord.
\newblock Think you have solved question answering? try arc, the ai2 reasoning challenge.
\newblock {\em arXiv:1803.05457v1}, 2018.

\bibitem{piqa}
Yonatan Bisk, Rowan Zellers, Jianfeng Gao, Yejin Choi, et~al.
\newblock Piqa: Reasoning about physical commonsense in natural language.
\newblock In {\em Proceedings of AAAI}, 2020.

\bibitem{sciq}
Johannes Welbl, Nelson~F. Liu, and Matt Gardner.
\newblock Crowdsourcing multiple choice science questions.
\newblock In {\em Proceedings of the 3rd Workshop on Noisy User-generated Text (ACL 2017)}, 2017.

\bibitem{boolq}
Christopher Clark, Kenton Lee, Ming-Wei Chang, Tom Kwiatkowski, Michael Collins, and Kristina Toutanova.
\newblock {B}ool{Q}: Exploring the surprising difficulty of natural yes/no questions.
\newblock In {\em Proceedings of NAACL-HLT}, 2019.

\bibitem{rein2024gpqa}
David Rein, Betty~Li Hou, Asa~Cooper Stickland, Jackson Petty, Richard~Yuanzhe Pang, Julien Dirani, Julian Michael, and Samuel~R. Bowman.
\newblock {GPQA}: A graduate-level google-proof q\&a benchmark.
\newblock In {\em First Conference on Language Modeling}, 2024.

\bibitem{amini2019mathqa}
Aida Amini, Saadia Gabriel, Peter Lin, Rik Koncel-Kedziorski, Yejin Choi, and Hannaneh Hajishirzi.
\newblock Mathqa: Towards interpretable math word problem solving with operation-based formalisms, 2019.

\bibitem{chen2021humaneval}
Mark Chen, Jerry Tworek, Heewoo Jun, Qiming Yuan, Henrique~Ponde de~Oliveira~Pinto, Jared Kaplan, Harri Edwards, Yuri Burda, Nicholas Joseph, Greg Brockman, Alex Ray, Raul Puri, Gretchen Krueger, Michael Petrov, Heidy Khlaaf, Girish Sastry, Pamela Mishkin, Brooke Chan, Scott Gray, Nick Ryder, Mikhail Pavlov, Alethea Power, Lukasz Kaiser, Mohammad Bavarian, Clemens Winter, Philippe Tillet, Felipe~Petroski Such, Dave Cummings, Matthias Plappert, Fotios Chantzis, Elizabeth Barnes, Ariel Herbert-Voss, William~Hebgen Guss, Alex Nichol, Alex Paino, Nikolas Tezak, Jie Tang, Igor Babuschkin, Suchir Balaji, Shantanu Jain, William Saunders, Christopher Hesse, Andrew~N. Carr, Jan Leike, Josh Achiam, Vedant Misra, Evan Morikawa, Alec Radford, Matthew Knight, Miles Brundage, Mira Murati, Katie Mayer, Peter Welinder, Bob McGrew, Dario Amodei, Sam McCandlish, Ilya Sutskever, and Wojciech Zaremba.
\newblock Evaluating large language models trained on code.
\newblock 2021.

\bibitem{austin2021mbpp}
Jacob Austin, Augustus Odena, Maxwell Nye, Maarten Bosma, Henryk Michalewski, David Dohan, Ellen Jiang, Carrie Cai, Michael Terry, Quoc Le, et~al.
\newblock Program synthesis with large language models.
\newblock {\em arXiv preprint arXiv:2108.07732}, 2021.

\bibitem{fairseq_dense}
Mikel Artetxe, Shruti Bhosale, Naman Goyal, Todor Mihaylov, Myle Ott, Sam Shleifer, Xi~Victoria Lin, Jingfei Du, Srinivasan Iyer, Ramakanth Pasunuru, et~al.
\newblock Efficient large scale language modeling with mixtures of experts.
\newblock In {\em Proceedings EMNLP}, 2022.

\bibitem{adamw}
Ilya Loshchilov and Frank Hutter.
\newblock Decoupled weight decay regularization.
\newblock In {\em Proceedings of ICLR}, 2019.

\bibitem{de2016spearman}
Joost~CF De~Winter, Samuel~D Gosling, and Jeff Potter.
\newblock Comparing the pearson and spearman correlation coefficients across distributions and sample sizes: A tutorial using simulations and empirical data.
\newblock {\em Psychological methods}, 21(3):273, 2016.

\end{thebibliography}
